\documentclass{article}
\usepackage{arxiv}
\usepackage[T1]{fontenc}
\usepackage[utf8]{inputenc}
\usepackage{amsmath,amssymb,amsthm}
\usepackage[round,authoryear]{natbib}
\usepackage[table]{xcolor}
\usepackage{float,booktabs,siunitx,xurl,graphicx,wrapfig,multirow,needspace}
\usepackage{capt-of}
\usepackage{microtype}
\usepackage{hyperref}
\usepackage{amsmath,amsfonts,bm}

\def\eqref#1{equation~\ref{#1}}
\def\1{\bm{1}}

\DeclareMathAlphabet{\mathsfit}{\encodingdefault}{\sfdefault}{m}{sl}
\SetMathAlphabet{\mathsfit}{bold}{\encodingdefault}{\sfdefault}{bx}{n}

\definecolor{wwlink}{HTML}{225F78}
\hypersetup{colorlinks=true,linkcolor=wwlink,citecolor=wwlink,urlcolor=wwlink,
  pdftitle={WorldWeave: Growing Persistent Geometric Worlds for Video Generation},
  pdfauthor={Yifan Huang, Lifan Jiang, Qingyue Hao, Cheng Chen, Boxi Wu, Xiaoxue Ren, Xiaofei He, Dehai Zhao}}
\renewcommand{\shorttitle}{\small WorldWeave}
\renewcommand{\headeright}{\small Preprint}
\renewcommand{\undertitle}{Preprint}
\newcommand{\methodname}{\textsc{WorldWeave}}
\newcommand{\wwhead}[1]{\begin{tabular}[c]{@{}c@{}}#1\end{tabular}}
\definecolor{wwgain}{RGB}{0,108,190}
\definecolor{wwrow}{gray}{0.92}
\definecolor{wwstudyteal}{RGB}{96,173,176}
\newcommand{\wwstudycell}[2]{\setlength{\unitlength}{1pt}\begin{picture}(19,24)\put(1,1){\color{wwstudyteal}\rule{4pt}{#2pt}}\put(11,20){\makebox(0,0){#1}}\end{picture}}
\newcommand{\wwyes}{\ensuremath{\checkmark}}
\makeatletter
\newcommand{\wwfinishwrap}{\par\ifnum\c@WF@wrappedlines>1\vspace{\dimexpr\baselineskip*\c@WF@wrappedlines-\baselineskip\relax}\fi\WFclear}
\renewcommand{\@maketitle}{%
  \begin{center}
  {\LARGE\bfseries\@title\par}\vspace{12pt}
  {\normalsize\normalfont\@author\par}\vspace{10pt}
  \end{center}\vspace{-5pt}\hrule\vspace{9pt}}
\makeatother
\title{WorldWeave: Growing Persistent\\Geometric Worlds for Video Generation}
\date{}
\author{Yifan Huang$^{1}$\thanks{Yifan Huang: \href{mailto:12621083@zju.edu.cn}{\texttt{12621083@zju.edu.cn}}.} \quad Lifan Jiang$^{1}$ \quad Qingyue Hao$^{1}$ \quad Cheng Chen$^{1}$\\[5pt]Boxi Wu$^{2}$ \quad Xiaoxue Ren$^{1}$ \quad Xiaofei He$^{1}$ \quad Dehai Zhao$^{1,\dagger}$\\[5pt]{\normalfont $^{1}$Zhejiang University \qquad $^{2}$Daerwen AI}\\[5pt]{\normalfont\small $^{\dagger}$Corresponding author.}}

\begin{document}

\maketitle

\begin{figure}[H]
\centering
\includegraphics[width=\linewidth]{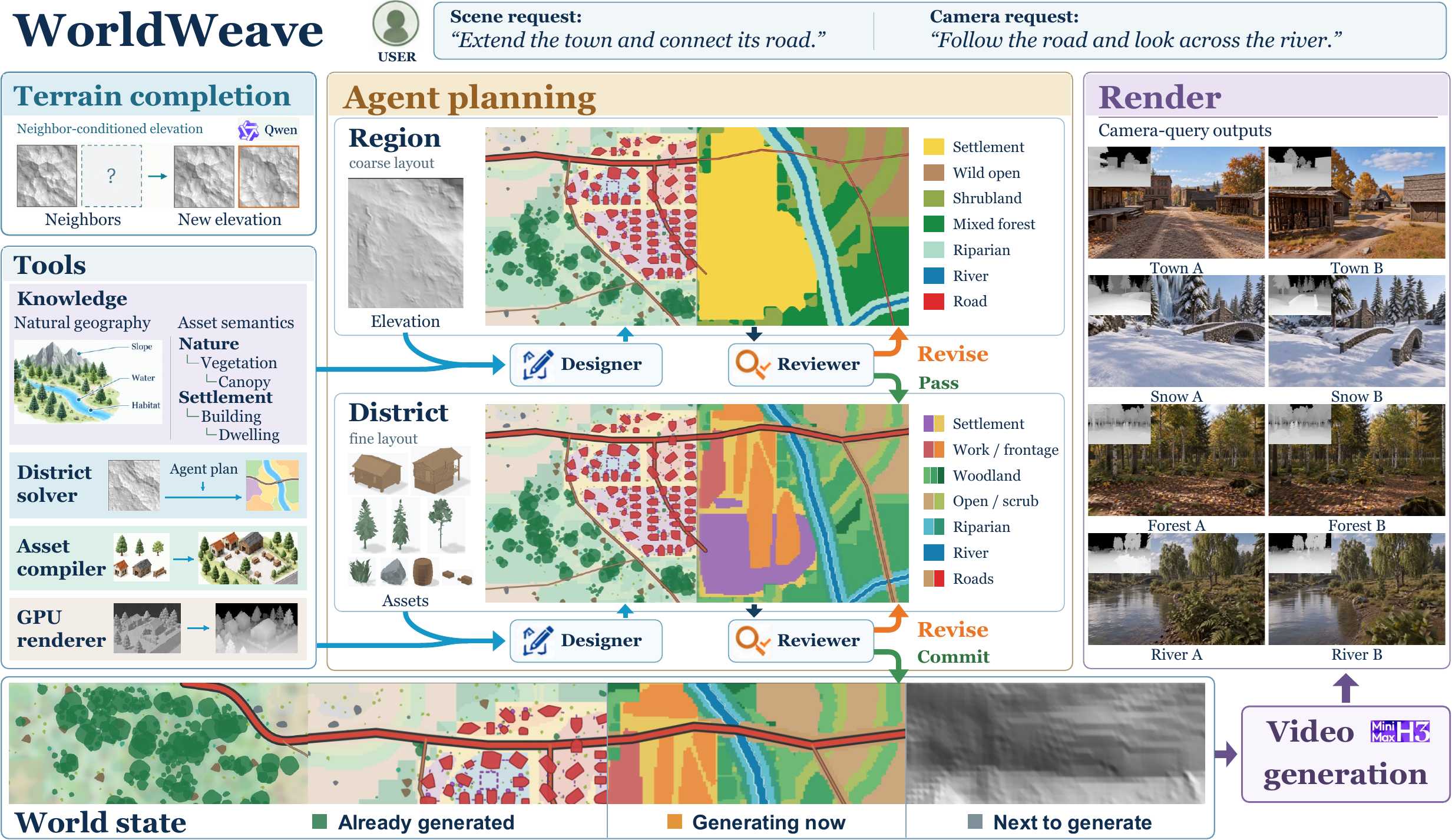}
\caption{\textbf{Overview of WorldWeave.} Terrain completion extends metric terrain, agent planning organizes and validates coarse-to-fine world layouts, and rendering queries the persistent world state to guide video generation. Previously generated regions are preserved while new regions are progressively added.}
\label{fig:overview}
\end{figure}

\begin{abstract}

Despite rapid progress, world models still lack explicit, persistent structural memory, making it difficult to preserve consistent world structure during continual scene expansion and cross-view revisits. To address this limitation, we present \methodname{}, a world generation framework that decouples world-state maintenance from visual rendering. Specifically, \methodname{} combines continual elevation-map generation with agent-guided scene organization and stitching to build an expandable explicit 3D world that incrementally extends structural memory while preserving existing structure. First, its terrain module uses diffusion-based image outpainting to generate continuous metric elevation maps under neighborhood conditioning and boundary constraints. Next, an agent integrates user intent, terrain evidence, and cross-region connectivity constraints to construct scenes through hierarchical semantic planning, deterministic geometry compilation, and local revision. Finally, during visual generation, planned camera trajectories query world geometry through a read-only interface, producing depth sequences that guide video synthesis without writing the generated results back into the world state. As a result, structural memory remains independent of short-window video generation, enabling continual expansion without predefined map boundaries and providing a consistent geometric basis for observations across trajectories and repeated visits.
\end{abstract}

\section{Introduction}

Generative world models have advanced rapidly in scene synthesis, view prediction and interactive environment generation, moving from local visual content toward explorable worlds~\citep{echo2026,zing2026,solarwm2026,sanawm2026}. Continual exploration requires more than realistic, temporally coherent observations: explored regions must persist outside the current view, new regions must connect to the existing environment, and revisits must encounter stable spatial structure. This requires structural memory beyond the current generation window, preserving world geometry, object identities and spatial relations as a common reference across time and viewpoints.

Despite improved visual generation, methods that rely primarily on images, video or finite observation histories still face challenges in maintaining explicit, persistent structural memory~\citep{worldmem2025,gen3c2025,evoke2026}. When observations implicitly carry world information, scene continuation depends on the generator re-inferring past content; local visual coherence does not directly establish lasting structural consistency. Expansion and revisitation expose two requirements: new regions must continue terrain, roads and waterways without altering established structure, and different observation trajectories must access the same world without reconstructing its geometric relations on each visit. A persistent world must therefore support both spatial addressing and relational composition: an extension must locate adjoining terrain, object support and continuing routes in a shared coordinate system. Persistence therefore governs both construction and observation.

We propose \methodname{}, a world generation framework that decouples world-state maintenance from visual rendering. An explicit 3D world in a shared metric coordinate system serves as structural memory, separating incremental construction from observation generation. The state stores terrain, regional organization, asset identities and transforms, and boundary interfaces inherited by later regions. Each expansion constructs an independent candidate from the committed state; only validated additions create a new version, leaving existing structural records unchanged. Visual generation reads a specified version, allowing different camera trajectories to reuse the same geometry. World structure thus persists independently of the current video window, while construction can continue without predefined map boundaries.

WorldWeave first uses diffusion-based image outpainting to generate complete metric terrain chunks from committed neighbors. A multiscale residual codec referenced to neighboring terrain maps elevation to an image representation; height and first-derivative constraints confined to the new chunk join it continuously to the existing surface. An agent then integrates user intent, terrain evidence and inherited interfaces to plan region roles, functional districts and asset relations. Deterministic solvers and geometry compilers realize these plans as spatial layouts and infrastructure connections. Program checks and matched-view geometry previews guide bounded local revision, after which accepted candidates are atomically committed as new world versions. Planned camera trajectories query this state for depth-conditioned video synthesis, with no RGB write-back.

This separation also defines what must remain consistent across observations: the scene geometry and relations belong to the world, while appearance is synthesized for each viewing request. New construction can therefore inherit off-screen structure without recovering it from previous RGB frames. The same distinction supports controlled evaluation: camera motion should reveal new content while retaining the organization of previously observed regions. We therefore assess structural consistency and revisit memory alongside camera compliance and visual quality, connecting preservation of scene organization with the ability to explore it. Figure~\ref{fig:overview} illustrates how scene requests and camera requests operate on this shared spatial reference.

Our contributions are threefold:
\begin{itemize}
  \item \textbf{Persistent structural-memory framework.} We decouple world-state maintenance from visual rendering, supporting incremental construction that preserves existing structure and shared geometric queries for observations across trajectories and revisits.
  \item \textbf{Continual metric terrain expansion.} We combine diffusion-based image outpainting, multiscale elevation-residual encoding, and new-side boundary constraints to generate complete metric terrain chunks and append them continuously.
  \item \textbf{Agent-guided scene construction.} We combine hierarchical semantic planning with deterministic geometry compilation, bounded local revision and validated commit to extend scenes following user intent, terrain conditions and inherited cross-region connections.
\end{itemize}

\section{Related Work}
\label{sec:related}

\paragraph{Video world models and structural memory.}
Echo-WM combines metric camera control with geometry conditioning~\citep{echo2026}; Zing-0.5 uses text, keyboard control and causal caching~\citep{zing2026}. SolarWM-5B uses camera-conditioned causal training~\citep{solarwm2026}, while SANA-WM uses hybrid linear attention for minute-long generation~\citep{sanawm2026}.

WorldMem retrieves frames by their states~\citep{worldmem2025}; ViewCrafter, WVD and Geometry-as-context condition view synthesis on geometry~\citep{viewcrafter2025,wvd2025,geometryascontext2026}. GEN3C unprojects previous observations into a 3D cache and renders it to guide generation~\citep{gen3c2025}, while WorldStereo and Lyra 2.0 connect views through geometric memory or correspondences~\citep{worldstereo2026,lyra22026}. Alaya-EVOKE-Turbo estimates geometry from generated frames and updates a camera-indexed world-state bank~\citep{evoke2026}; AlayaWorld combines a 3D cache with compressed frame history~\citep{alayaworld2026}. WorldWeave instead maintains memory through validated scene construction, with video generation reading committed geometry.

\paragraph{Persistent and expandable 3D worlds.}
Persistent Nature uses an extendable layout grid and camera-independent decoder~\citep{persistentnature2023}. Text2Room, WonderJourney and WonderWorld grow scenes through view synthesis and geometric integration~\citep{text2room2023,wonderjourney2024,wonderworld2025}; ScenePainter tracks concept relations to limit semantic drift~\citep{scenepainter2025}, and One2Scene establishes a shared Gaussian scaffold before generating novel views~\citep{one2scene2026}. InfiniCube follows a construction-first approach, expanding semantic voxel worlds from HD maps and vehicle boxes and rendering guidance buffers for driving videos~\citep{infinicube2025}. XCube and LT3SD use sparse voxels and latent trees~\citep{xcube2024,lt3sd2025}; BlockFusion, SceneFactor and NuiScene extend spatial blocks~\citep{blockfusion2024,scenefactor2025,nuiscene2025}. We preserve committed structure and inherit terrain and scene interfaces.

\paragraph{Agent-guided incremental scene construction.}
Infinigen supplies procedural environments~\citep{infinigen2023,infinigenindoors2024}; 3D-GPT, SceneCraft and SceneX expose construction through language-driven programs~\citep{3dgpt2023,scenecraft2024,scenex2024}. Holodeck supports language- and vision-guided layouts~\citep{holodeck2024,holodeck22025}; Scenethesis combines visual guidance with geometric constraints, while scene graphs and hierarchical motifs organize relations~\citep{scenethesis2026,control3dscene2025,hsm2026}. WorldClaw builds region-aware terrain and places editable assets with rendering feedback~\citep{worldclaw2026}. WorldGen imposes navigational structure to obtain traversable worlds~\citep{worldgen2026}; Code2Worlds separates object generation from environmental orchestration and refines scene programs through feedback~\citep{code2worlds2026}. Our planning inherits neighboring interfaces and uses deterministic compilation and bounded revision before committing additions.

\paragraph{Terrain generation and boundary continuity.}
Hydrology-based modeling organizes terrain around drainage~\citep{hydrology2013}. Learned approaches include sketch-conditioned diffusion with upscaling~\citep{lochner2023terrain}, joint height--texture synthesis with separate latent encoders~\citep{terrafusion2025} and text-conditioned generation trained on geospatial data~\citep{mesa2025}. InfiniteDiffusion supports seed-consistent random access through overlapping queries~\citep{infinitediffusion2026}. Our fixed-order expansion instead generates metric chunks against committed neighbors and joins them through new-side height and derivative constraints.

\section{Method}
\label{sec:method}

\begin{figure}[!t]
\centering
\includegraphics[width=\linewidth]{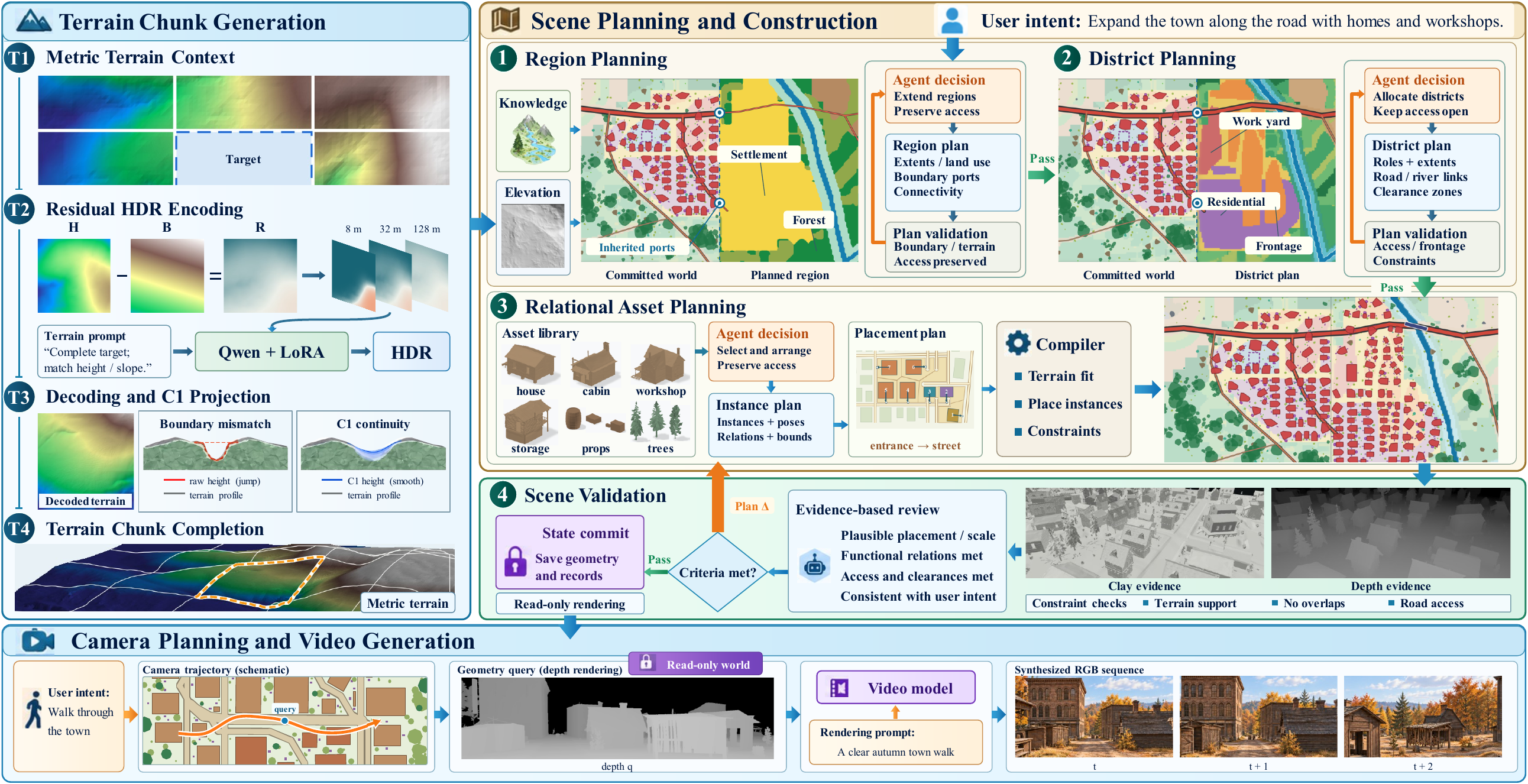}
\caption{\textbf{WorldWeave pipeline.} Neighbor-conditioned terrain generation uses residual HDR encoding and new-side joining. The agent plans regions, districts and asset relations, while deterministic tools compile geometry and provide structural and visual evidence for revision. Validated additions extend persistent world state; camera trajectories query its geometry to condition video generation without writing RGB back into the world.}
\label{fig:pipeline}
\end{figure}

\subsection{Persistent Structural Memory}
\label{sec:world_state}

WorldWeave starts from a seed terrain chunk and maintains an explicit world in a shared metric coordinate system. This structural memory has committed version $M_k=(\mathcal{T}_k,\mathcal{P}_k,\mathcal{I}_k,\mathcal{J}_k)$ after $k$ accepted additions: $\mathcal{T}_k$ stores elevation and boundary derivatives, $\mathcal{P}_k$ stores regional organization and object relations, $\mathcal{I}_k$ stores asset mesh references, identities and world transforms, and $\mathcal{J}_k$ stores exposed boundary interfaces. These interfaces specify terrain attachment and the positions and types of road or water connections inherited by new regions.

An expansion request specifies intent $u$ and frontier chunk $q$. Construction reads $M_k$ and forms an independent candidate $\Delta M_q$ through terrain expansion and hierarchical scene compilation (Figure~\ref{fig:pipeline}), using the committed terrain and inherited interfaces. Validation and commit (Section~\ref{sec:commit}) determine whether the candidate becomes a new version. A camera query selects an existing committed version; changing the observation does not change its structural records. The state links semantic and geometric descriptions through stable instance identities. Regional plans specify which parts of the terrain serve each function, while instance transforms place the corresponding assets in the same metric frame. Exposed interfaces transfer this organization across chunk boundaries: a new region receives both the neighboring surface and the road or water connections it must extend. This gives local construction a consistent global reference as the map grows.

\subsection{Continual Metric Terrain Expansion}
\label{sec:terrain}

A candidate terrain chunk contains $512\times512$ elevation samples at 0.5\,m spacing, giving a nominal $256\times256$\,m footprint. Its context comprises one to three committed axial neighbors: one side, two opposite sides, two adjacent sides, or three sides. We generate a complete target in one image-outpainting call and repeat this operation in a fixed expansion order, using previously committed outputs as subsequent context.

To encode metric height within the image model's dynamic range, we fit a reference surface $b_q(x,y)$ from valid neighboring boundary bands. The unknown target does not enter this fit. For elevation $h_q$, the residual $r_q$ is encoded through three monotone channels:
\begin{equation}
 r_q=h_q-b_q,\qquad
 c_{q,j}=\tfrac12+\tfrac12\tanh(r_q/s_j),\qquad
 (s_1,s_2,s_3)=(8,32,128)\,\mathrm{m}.
 \label{eq:terrain_codec}
\end{equation}
Each encoded sample is repeated in a $2\times2$ image block before VAE encoding, adding redundancy without changing the metric sampling grid. The reference surface carries the elevation level and broad trend supplied by the context, leaving the residual channels to describe local relief. Small channel scales allocate greater sensitivity to subtle variations, whereas larger scales retain information over stronger relief. The same reference is restored after decoding, so neighboring chunks remain expressed in world units even though generation takes place in image space.

We adapt Qwen-Image-Edit-2509~\citep{qwenimage2025} with rank-32 LoRA~\citep{lora2021}, freezing its VAE and text conditioner. Encoded neighbors and a fixed completion prompt condition generation. We arrange the context on a spatial canvas, retaining neighbor directions around a neutral target slot. This exposes opposite or adjacent boundary conditions in one layout, allowing the target to be completed jointly against all attached sides. Let $\mathcal{V}_q$ index valid target loss locations, $\ell_q(x)$ denote the local diffusion loss, and $d_q(x)$ the metric distance to the nearest attached edge. We use
\begin{equation}
\begin{aligned}
 w_q(x)&=1+\tfrac32\!\left[1+\cos\!\left(\pi\min\{d_q(x)/a,1\}\right)\right],\\
 \mathcal{L}_q&=\frac{\sum_{x\in\mathcal{V}_q}w_q(x)\,\ell_q(x)}
 {\sum_{x\in\mathcal{V}_q}w_q(x)},\qquad a=16\,\mathrm{m}.
\end{aligned}
\label{eq:edge_loss}
\end{equation}
Weights are evaluated on the loss grid. The nearest-edge distance implements the maximum over intersecting edge bands; normalization keeps loss scales comparable across targets.

After VAE decoding, each channel gives a residual estimate $\hat r_{q,j}=s_j\operatorname{atanh}(2\hat c_{q,j}-1)$, with channel values clipped inside $(0,1)$. We combine the repeated estimates using inverse-sensitivity weighting, downweighting near-saturated channels, and refine the result with five robust projection iterations onto the three-channel encoding curve. Adding $b_q$ recovers the metric prediction $\hat h_q=b_q+\hat r_q$ on the original sampling grid. For each attached side $e\in\mathcal{E}_q$, the neighbor supplies height $g_e$ and directional derivative $v_e$ on the world-coordinate interface $\Gamma_e$, using a shared normal $n_e$. The joined surface satisfies
\begin{equation}
\begin{aligned}
 h_q^+&=\hat h_q+\delta_q,\qquad \operatorname{supp}(\delta_q)\subseteq \mathcal{B}_q,\\
 h_q^+\big|_{\Gamma_e}&=g_e,\qquad
 \partial_{n_e}h_q^+\big|_{\Gamma_e}=v_e,\qquad e\in\mathcal{E}_q .
\end{aligned}
\label{eq:seam_conditions}
\end{equation}
\begin{wrapfigure}{R}{.48\linewidth}
\centering
\includegraphics[width=\linewidth]{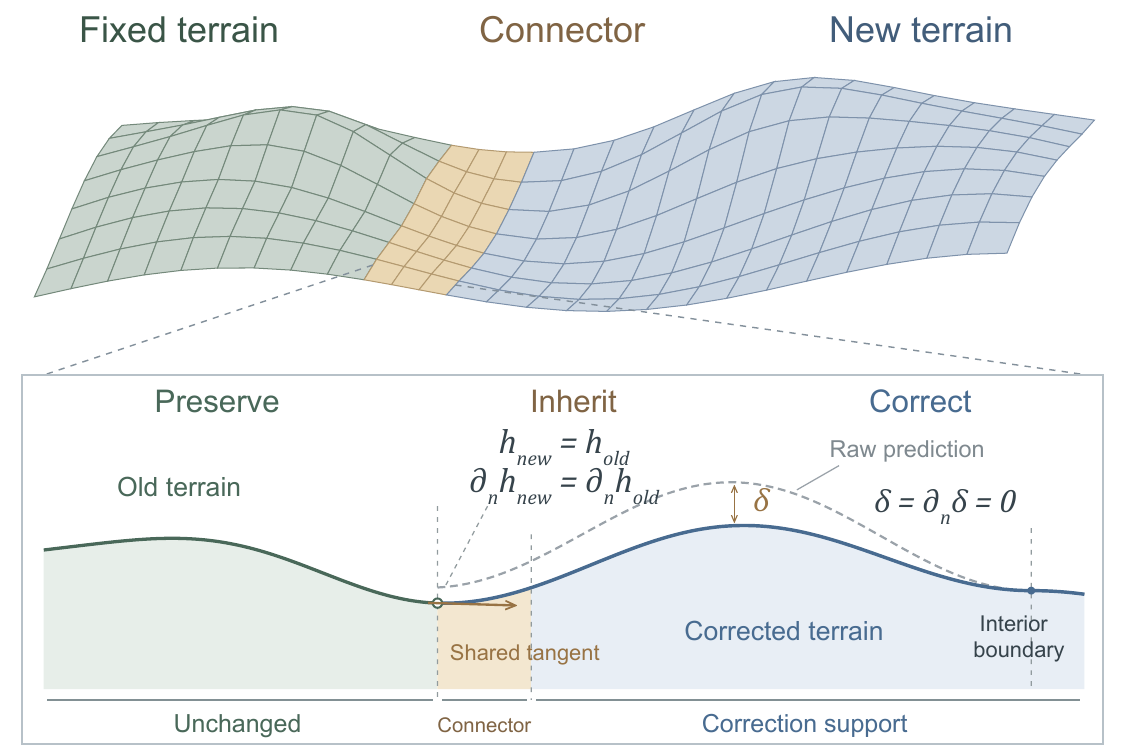}
\caption{New-side terrain joining. Old samples remain fixed; newly owned connector cells inherit boundary height and normal derivative. The correction vanishes at the interior support boundary. Schematic.}
\label{fig:joining_conditions}
\end{wrapfigure}
Here $\mathcal{B}_q$ includes the new-side height-transition band and connector cells between existing sample-center grids. New connector cells are bicubic Hermite patches whose old-facing edges inherit boundary heights and derivatives (Figure~\ref{fig:joining_conditions}); slope correction uses an independent taper. Corrections vanish at their inner support boundaries, with $\delta_q=\partial_n\delta_q=0$. Committed geometry is preserved, and modification magnitude is reported separately from continuity (Appendix~\ref{app:terrain_details}). Connector cells bridge the gap between the old and new sample-center grids, providing a shared boundary for subsequent terrain and scene construction. Height correction adjusts the offset between chunks, whereas slope correction controls the approach to the inherited boundary slope. Separate support widths let the surface satisfy both conditions while confining derivative-induced displacement near the seam.

\subsection{Hierarchical Scene Planning and Compilation}
\label{sec:planning}

Deterministic terrain analysis extracts slope, local relief, drainage and buildable-area evidence from the accepted elevation field. The agent combines this evidence with $u$, inherited interfaces $\mathcal{J}_k$, and available asset capabilities to assign broad region roles, such as settlement, woodland or river corridor. A region solver converts these semantic assignments into spatial extents. After region review (Section~\ref{sec:commit}), the agent specifies district roles, adjacency and access requirements. A deterministic solver resolves their extents and infrastructure under inherited road and water interfaces, producing functional or ecological subdivisions with the access, clearance and terrain conditions required by downstream assets.

Within each district, the agent turns its functional requirements into an asset-relation plan. The plan assigns primary, companion and service roles, specifies shared spaces and orientation, and connects buildings and vegetation groups to the district's roads and ecological zones. The typed catalog supplies each asset's dimensions, support anchors, entrance direction and required context. These attributes connect semantic choices to the spatial constraints inherited from regional and district planning. The hierarchy passes explicit commitments between scales: region boundaries delimit admissible land use, district connections reserve routes through those regions, and asset relations specify how individual instances participate in that organization. The deterministic compiler resolves the plan into stable instance identities and metric transforms under the spatial and asset contracts in Appendix~\ref{app:scene_details}. It fits assets to supporting surfaces, realizes infrastructure connections and vegetation groupings, and preserves access corridors and collision clearance. For $n_q$ candidate instances with required geometric relations $\mathcal R_q$, acceptance requires
\begin{equation}
 \mathcal I_q=\{(a_i,T_i,\rho_i,d_i)\}_{i=1}^{n_q},\qquad
 g_r(\mathcal I_q,h_q^+;M_k)=1\quad(r\in\mathcal R_q).
 \label{eq:compiled_scene_relations}
\end{equation}
Here $a_i$ identifies catalog geometry, $T_i$ its world transform, $\rho_i$ its semantic role and $d_i$ its district. Each predicate $g_r$ checks a catalog-supported relation against the candidate and committed context. Failed predicates retain instance and relation identifiers, locating the plan entries to revise. The plan, geometry and checks form the scene portion of $\Delta M_q$.

\subsection{Bounded Revision and Validated Commit}
\label{sec:commit}

Region review checks inherited connections, coverage and terrain feasibility before district refinement. After compilation, program checks evaluate support, collision and connectivity in parallel with matched-view GPU geometry and depth previews. The tests establish geometric validity; matched viewpoints show whether the compiled arrangement realizes the intended regional organization. The agent receives both for the same candidate, allowing a revision to address the responsible functional area or obstructed connection while retaining unaffected content. Accepted regions then expose their outer interfaces as context for the next expansion, linking each local planning cycle to continued world construction. Review evidence is bound to the candidate revision: after an edit, program reports and matched-view previews are regenerated for that revision. This keeps acceptance tied to the geometry that will be committed.

Let $\pi_q^{(t)}$ be the candidate plan after $t$ revisions and $K$ the revision budget. A failed review returns a structured delta $\delta\pi_q^{(t)}$ targeting the responsible region, district or asset relation:
\begin{equation}
\pi_q^{(t+1)}=\operatorname{Edit}(\pi_q^{(t)},\delta\pi_q^{(t)}),\qquad
\Delta M_q^{(t+1)}=\operatorname{Compile}(\pi_q^{(t+1)};M_k),\qquad t<K.
\label{eq:local_revision}
\end{equation}
$\operatorname{Edit}$ updates the indicated plan entries; $\operatorname{Compile}$ recomputes affected geometry with $M_k$ read-only. Reviews use fixed acceptance criteria and reject candidates that exhaust the revision budget.

When all geometric checks and agent reviews pass, the candidate is atomically published as a new world version. Let $\Omega_k$ denote the terrain, semantic-plan, instance and boundary-geometry records already committed in $M_k$. The update appends accepted records while preserving those fields:
\begin{equation}
 M_{k+1}=M_k\oplus\Delta M_q,\qquad
 M_{k+1}\big|_{\Omega_k}=M_k\big|_{\Omega_k}.
 \label{eq:world_commit}
\end{equation}
Here $\oplus$ denotes the validated addition, including new exposed interfaces for subsequent expansion. The equality preserves the terrain, organization and instance records reused by future expansions and observations. Visibility is computed from the complete geometry of the selected version.

\subsection{Read-only Queries for Video Generation}
\label{sec:video_query}

Observation generation selects a committed world version independently of expansion. A viewing request specifies desired motion and targets, either supplied directly by the user or sampled by the agent from the user's intent. The camera planner converts this request into intrinsics, poses and timing $C_{1:T}$, then checks the trajectory for clearance and the required viewpoints. GPU queries resolve nearest visible intersections across terrain and asset meshes, returning metric depth with its convention, validity mask and calibration. Per-pixel hit identities associate queried surfaces with persistent asset records. They bind named camera targets to world geometry when assessing visibility along a trajectory. The resulting RGB video visualizes this world version.

As the camera moves, visible surfaces are recomputed from this common geometry, providing spatially coordinated conditions throughout the trajectory. For video generator $F$, let $A_F$ adapt depths from query $Q$. Its supported optional style input $s_F$ comprises text, reference images, or both:
\begin{equation}
 D_{1:T}=Q(M_k,C_{1:T}),\qquad
 V_{1:T}=F\bigl(A_F(D_{1:T}),s_F\bigr).
 \label{eq:video_readout}
\end{equation}
A surface point $X$ retains its world coordinates in the selected version. For two views in which it is visible, the homogeneous image coordinates obey
\begin{equation}
\widetilde x_t(X)\sim K_tR_t^\top(X-c_t),\qquad
 \widetilde x_s(X)\sim K_sR_s^\top(X-c_s).
\label{eq:shared_world_projection}
\end{equation}
Here $K_t$, $R_t$ and $c_t$ denote intrinsics, camera-to-world rotation and camera center. Both projections refer to the same terrain or asset surface, supplying a shared geometric correspondence through viewpoint changes and revisits. The depth sequence encodes how camera motion changes visibility, including occlusions and the boundaries of terrain and assets.

The adapter converts metric depth to the generator's required encoding, resolution and frame rate while preserving camera calibration. Available style references control appearance, while depth supplies the spatial structure along the requested trajectory. Generated RGB never writes back into world state. Appendix~\ref{app:query_details} specifies revision, publication and observation.
\begin{figure}[!t]
\centering
\includegraphics[width=\linewidth]{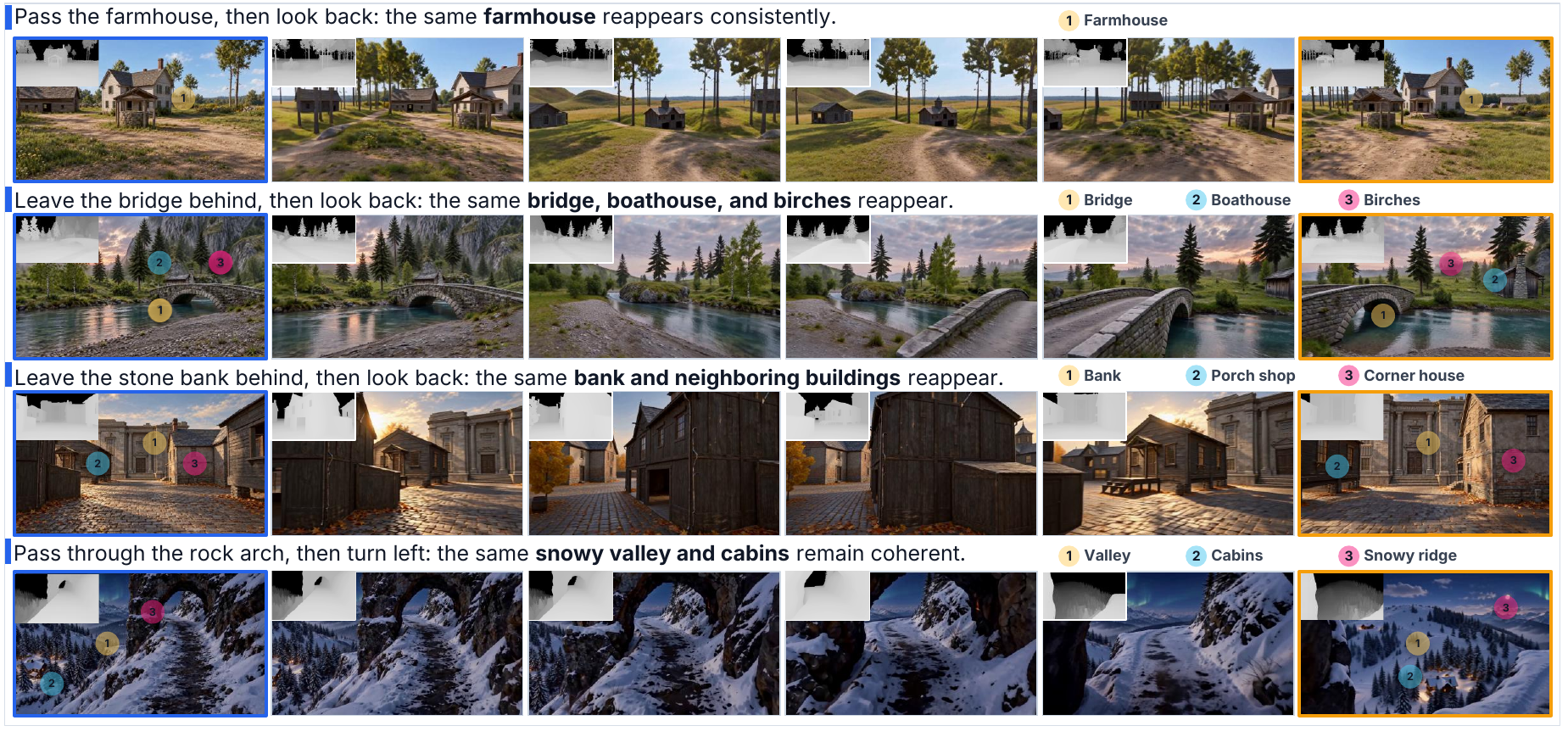}
\caption{World-grounded video generation across four scenes. Each row follows a viewing trajectory from left to right, showing revisitation or continued exploration. Insets show depth guidance, and numbered markers identify corresponding scene elements across views.}
\label{fig:demo_main}
\end{figure}

\section{Experiments}
\label{sec:experiments}

\subsection{Experimental Setup}
\label{sec:experiment_setup}

\paragraph{Benchmark.}
We evaluate 120 tasks covering forward exploration, viewpoint changes, and occlusion or off-screen return. Terrain relief, vegetation biomes (e.g., forests and grasslands), and settlement layouts (e.g., villages and towns) vary across scenes. Seasonal appearance and lighting are varied as rendering attributes. Videos are evaluated over 15 seconds at 12 fps.

\paragraph{Baselines.}
We compare with six open-source world models: SANA-WM, Zing, SolarWM-5B, AlayaWorld, Alaya-EVOKE-Turbo and JoyAI-Echo-1.5 (WM)~\citep{sanawm2026,zing2026,solarwm2026,alayaworld2026,evoke2026,echo2026}, and four video baselines: Seedance 2.0 (closed-source), Wan3.0, Kling 3.0 and our base generator MiniMax-H3. Since world models commonly require a first frame and camera controls, we provide them with the same first frame as the WorldWeave-generated video and the camera trajectory used to produce its depth input. All models receive the same task and scene-description prompt. WorldWeave additionally receives depth queried from the constructed world.

\paragraph{Metrics.}
We assess visual quality with IQ and AQ from VBench~\citep{vbench2024}, and revisit memory and camera compliance with SMC and Cam. We jointly calibrate smoothness and geometric consistency for realized motion, yielding MN-MS, MC-GeCo~\citep{geco2025}, MC-MEt3R~\citep{met3r2025} and MC-GeoCon~\citep{geocon2026}, to reduce systematic scoring biases across different camera-motion speeds. Appendix~\ref{app:evaluation} defines these metrics and their motion calibration.

\begin{table}[t]
\caption{World-video comparison. The first six methods are open-source world models; the remaining baselines are video models. Bold/underline indicate best/second-best scores; yellow highlights the three largest relative gains over MiniMax-H3 (Base), with signed percentage changes.}
\label{tab:main_results}
\centering
\fontsize{8}{9.5}\selectfont
\setlength{\tabcolsep}{.8pt}
\begin{tabular}{@{}ccccccccc@{}}
\toprule
\multirow{2}{*}[-6.9pt]{Method} & \multicolumn{2}{c}{Visual quality} & \multicolumn{2}{c}{Memory and camera control} & \multicolumn{4}{c}{Structural consistency}\\
\cmidrule(lr){2-3}\cmidrule(lr){4-5}\cmidrule(l){6-9}
& \wwhead{Imaging\\quality\ensuremath{\uparrow}} & \wwhead{Aesthetic\\quality\ensuremath{\uparrow}} &
\wwhead{Structural\\memory\ensuremath{\uparrow}} & \wwhead{Camera\\compliance\ensuremath{\uparrow}} &
MN-MS\ensuremath{\uparrow} & MC-GeCo\ensuremath{\downarrow} &
MC-MEt3R\ensuremath{\downarrow} & MC-GeoCon\ensuremath{\downarrow}\\
\midrule
SANA-WM & 0.739 & 0.619 & 0.328 & 55.5 & 0.973 & 0.113 & 0.203 & 0.189\\
Zing & 0.763 & 0.685 & 0.989 & 55.8 & 0.974 & 0.0698 & {\underline{0.129}} & 0.142\\
SolarWM-5B & 0.755 & 0.604 & 0.370 & 54.7 & 0.956 & 0.0786 & 0.166 & 0.217\\
AlayaWorld & 0.784 & 0.641 & 1.51 & 56.4 & 0.896 & 0.926 & 0.239 & 0.295\\
EVOKE-Turbo & 0.784 & \textbf{0.718} & 1.12 & 61.6 & 0.961 & 0.105 & 0.182 & 0.249\\
Echo-WM & \textbf{0.791} & 0.655 & {\underline{2.21}} & 63.5 & 0.968 & {\underline{0.0604}} & 0.135 & 0.135\\

\midrule
Seedance 2.0 & 0.717 & 0.617 & 2.14 & {\underline{69.8}} & {\underline{0.980}} & 0.101 & 0.170 & 0.307\\
Wan3.0 & {\underline{0.786}} & 0.702 & 1.63 & 59.1 & 0.938 & 0.146 & 0.184 & {\underline{0.131}}\\
Kling 3.0 & 0.739 & 0.588 & 2.09 & 59.3 & 0.978 & 0.0745 & 0.155 & 0.163\\
MiniMax-H3 (\textit{Base}) & 0.725 & 0.617 & 0.438 & 50.8 & 0.979 & 0.0633 & 0.140 & 0.178\\
\rowcolor{wwrow}
\textbf{WorldWeave} & 0.781 & {\underline{0.704}} & \cellcolor{yellow!70}\textbf{2.30}\,(+426.7\%) & \cellcolor{yellow!70}\textbf{72.3}\,(+42.2\%) & \textbf{0.981} & \textbf{0.0573} & \textbf{0.122} & \cellcolor{yellow!70}\textbf{0.121}\,(-32.2\%)\\
\bottomrule
\end{tabular}
\end{table}

\subsection{Main Quantitative Results}
\label{sec:main_results}

Table~\ref{tab:main_results} compares visual quality, structural consistency and camera control. WorldWeave ranks fifth in IQ and second in AQ, while leading MN-MS, MC-GeCo, SMC, Cam, MC-MEt3R and MC-GeoCon. Its strongest gains therefore concern maintaining an explorable world: even with greater mean image-space motion than six comparison methods, WorldWeave retains leading structural-consistency and camera-compliance scores alongside competitive appearance and smooth motion. Figure~\ref{fig:demo_main} illustrates the corresponding video sequences across revisitation and exploration tasks.

The Base comparison examines the benefit of world-grounded conditioning within the same video-model family. Relative to MiniMax-H3, WorldWeave improves all eight measures. The largest gains are in SMC (+426.7\%) and Cam (+42.2\%), together with a 32.2\% reduction in MC-GeoCon error. IQ and AQ also improve, supporting both spatial organization and appearance.

\subsection{Structural Consistency and Camera Compliance}
\label{sec:structural_results}

\paragraph{Geometric consistency.}
WorldWeave achieves the lowest MC-GeCo and MC-MEt3R errors, 0.0573 and 0.122, improving over MiniMax-H3 by 9.6\% and 12.5\%, respectively. The lower motion-calibrated errors indicate more coherent geometry and reprojected features across generated views. MC-GeoCon corroborates this result through geometric correspondences. These results support transferring spatial organization from shared depth guidance into generated video.

\paragraph{Memory across revisits.}
WorldWeave obtains the highest SMC score, 2.30 compared with 2.21 for the runner-up Echo-WM, while MiniMax-H3 scores 0.438. This improvement strengthens memory over disappearance and return. When a target leaves the view or becomes occluded, later observations must recover its identity and spatial relations. Reusing a committed world version provides the same target geometry and surrounding layout on return, reducing the need to reconstruct these relations from a limited visual history. Appendix~\ref{app:comparisons} provides matched occlusion and return sequences in farmstead, town and woodland scenes.

\paragraph{Camera compliance.}
WorldWeave also ranks first in Cam at 72.3, ahead of Seedance 2.0 by 2.49 points. Relative to MiniMax-H3's 50.8, this is a 42.2\% increase in score, reflecting stronger adherence to the requested translations, turns and returns. On the 120 matched tasks, its mean optical-flow amplitude exceeds those of SANA-WM, EVOKE-Turbo, AlayaWorld, Seedance 2.0, Wan3.0 and Kling 3.0 by 8.1--24.1\%. Even with this greater realized motion, WorldWeave maintains leading structural-consistency scores, supporting exploration with coherent changes in viewpoint. Appendix~\ref{app:corrections} contrasts favorable raw scores with texture degradation or limited motion.

\subsection{Ablation Studies}
\label{sec:ablations}
\begin{wrapfigure}{l}{.44\linewidth}
\centering
{\small Canvas\par}
\includegraphics[width=\linewidth]{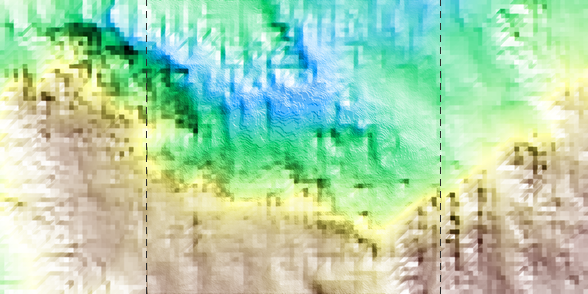}
{\small Multi-image\par}
\includegraphics[width=\linewidth]{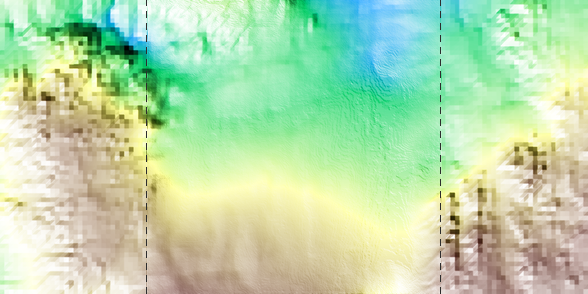}
\caption{Opposite-neighbor completion with canvas (top) and multi-image inputs (bottom). Additional cases: Appendix~\ref{app:terrain_visual}.}
\label{fig:terrain_input_main}
\end{wrapfigure}

We evaluate elevation encoding, terrain conditioning and joining, followed by planning-model substitution and video-generation seed variation (Appendix~\ref{app:ablations}).

\paragraph{Metric elevation encoding.}
Using residual rather than absolute grayscale reduces mean VAE round-trip RMSE from 2.182 to 0.469\,m (Table~\ref{tab:terrain_codec_results}(a)). Multiscale residual channels further reduce it to 0.158\,m, and the full codec reaches 0.136\,m, a further 14.0\% reduction. Referencing neighboring terrain removes the large elevation offset, while the channel scales retain sensitivity across different relief levels. These reductions demonstrate improved elevation-codec fidelity: context referencing and multiscale channels jointly preserve metric height through the frozen image VAE.

\begin{table}[t]
\caption{Terrain ablations of (a) elevation encoding, (b) input organization and (c) terrain joining. $\Delta h$ denotes whole-target RMS modification.}
\label{tab:terrain_codec_results}
\label{tab:terrain_input_results}
\label{tab:terrain_join_results}
\centering\fontsize{6}{7.2}\selectfont
\setlength{\tabcolsep}{1pt}
\begin{tabular}[t]{@{}c@{}}
\textbf{(a) Elevation encoding}\\[2pt]
\begin{tabular}{@{}cccc@{}}
\toprule
\wwhead{Reference\\subtraction} & \wwhead{Multiscale\\HDR} & \wwhead{Spatial\\repetition} & RMSE (m)\ensuremath{\downarrow}\\
\midrule
-- & -- & -- & 2.182\\
-- & \wwyes & -- & 1.935\\
\wwyes & -- & -- & 0.469\\
\wwyes & \wwyes & -- & 0.158\\
\wwyes & \wwyes & \wwyes & \textbf{0.136}\\
\bottomrule
\end{tabular}\\
\end{tabular}\hfill%
\begin{tabular}[t]{@{}c@{}}
\textbf{(b) Terrain input}\\[2pt]
\begin{tabular}{@{}ccccc@{}}
\toprule
\multirow{2}{*}{Input}&\multicolumn{2}{c}{Height gap (m)\ensuremath{\downarrow}}&\multicolumn{2}{c}{Slope gap (m/m)\ensuremath{\downarrow}}\\
\cmidrule(lr){2-3}\cmidrule(l){4-5}
& Multi & Canvas & Multi & Canvas\\
\midrule
One & 0.763 & \textbf{0.264} & \textbf{0.240} & 0.278\\[.595pt]
Opposite & 1.394 & \textbf{0.333} & 0.373 & \textbf{0.338}\\[.595pt]
L-shaped & 1.076 & \textbf{0.296} & 0.310 & \textbf{0.275}\\[.595pt]
Three & 1.214 & \textbf{0.368} & 0.361 & \textbf{0.328}\\[.595pt]
\bottomrule
\end{tabular}\\
\end{tabular}\hfill%
\begin{tabular}[t]{@{}c@{}}
\textbf{(c) Terrain joining}\\[2pt]
\begin{tabular}{@{}cccccc@{}}
\toprule
\wwhead{Height\\correction}&\wwhead{Derivative\\inheritance}&\wwhead{Slope\\taper}&$H_c\downarrow$ (m)&$S_c\downarrow$ (m/m)&$\Delta h$ (m)\\
\midrule
--&--&--&0.319&0.638&0\\[1.8pt]
\wwyes&--&--&0&0.359&0.0773\\[1.8pt]
\wwyes&\wwyes&--&0&0&0.2088\\[1.8pt]
\wwyes&\wwyes&\wwyes&0&0&\textbf{0.0825}\\[1.8pt]
\bottomrule
\end{tabular}\\
\end{tabular}
\end{table}

\noindent\textbf{Spatially arranged terrain conditioning.}
Canvas conditioning yields lower boundary-height gaps in all four neighbor configurations (Table~\ref{tab:terrain_input_results}(b)), reducing them by 65.4--76.1\% relative to independent multi-image references. The remaining gaps are 0.264--0.368\,m for nominally 256\,m chunks. Slope continuity also improves in the two- and three-neighbor settings, although the single-neighbor case favors multi-image input. This pattern supports arranging neighboring terrain in its spatial context, especially when several boundaries must be satisfied together. Figure~\ref{fig:terrain_input_main} shows the opposing-neighbor case; Appendix~\ref{app:terrain_visual} compares all configurations. Training configurations and evaluation details are specified in Appendix~\ref{app:ablation}.

\paragraph{Continuous terrain joining.}
Table~\ref{tab:terrain_join_results}(c) compares four settings on 256 matched targets. Height-only joining closes the height gap but leaves slope residual 0.359; full derivative inheritance reduces both residuals below numerical tolerance. Independent slope taper preserves continuity while reducing mean whole-target modification from 0.2088 to 0.0825\,m (60.5\%). All joining variants connect every target and preserve valid existing heights and derivatives, supporting separate height and slope transition widths (Appendix~\ref{app:joining_details}).

\paragraph{Planning model.}
Replacing the planner changes settlement density and infrastructure while retaining the shared terrain and cross-region organization. Both planners continue inherited roads: the alternative favors woodland and small settlement groups, while the default extends a denser street network. Appendix~\ref{app:planner_comparison} compares full layouts with identical terrain and the same existing region.

\paragraph{Random-seed sensitivity.}
Across five fixed scene--trajectory pairs and five generation seeds, the seed-wise mean IQ, AQ and MN-MS have coefficients of variation below 0.4\%; those of the three geometric metrics range from 3.1\% to 6.3\%. Thus, the aggregate computational measures remain comparatively stable as the noise realization changes. Event-based SMC and Cam vary more strongly. Appendix~\ref{app:seed_stability} provides per-task scores, seed-wise means and dispersion, including the observed black-sky case under fixed geometry, first frame, prompt and depth.

\wwfinishwrap
\Needspace{150pt}
\subsection{User Study}
\label{sec:user_study}
\par\noindent\begin{minipage}[t]{.48\linewidth}\vspace{0pt}
\centering
\fontsize{7}{8}\selectfont
\setlength{\tabcolsep}{.8pt}
\begin{tabular}{@{}c|ccccccc@{}}
& \wwhead{SANA\\WM} & Zing & \wwhead{Solar\\WM} & \wwhead{Alaya\\World} & \wwhead{EVOKE\\Turbo} & \wwhead{Echo\\WM} & \textbf{Ours}\\
\hline
\wwhead{Camera\\consistency}
& \wwstudycell{2.65}{8.48} & \wwstudycell{2.76}{8.84} & \wwstudycell{1.99}{6.36} & \wwstudycell{3.23}{10.35} & \wwstudycell{2.84}{9.08} & \wwstudycell{3.92}{12.53} & \wwstudycell{\textbf{4.57}}{14.61}\\
\wwhead{Object\\consistency}
& \wwstudycell{1.52}{4.85} & \wwstudycell{2.88}{9.21} & \wwstudycell{1.49}{4.77} & \wwstudycell{3.06}{9.80} & \wwstudycell{2.52}{8.05} & \wwstudycell{3.42}{10.95} & \wwstudycell{\textbf{4.73}}{15.15}\\
\wwhead{Structure/\\texture}
& \wwstudycell{1.61}{5.16} & \wwstudycell{2.85}{9.11} & \wwstudycell{1.84}{5.89} & \wwstudycell{2.68}{8.57} & \wwstudycell{2.64}{8.44} & \wwstudycell{3.32}{10.63} & \wwstudycell{\textbf{4.52}}{14.47}\\
\end{tabular}
\captionof{table}{User ratings (1--5; higher is better).}
\label{tab:user_study}
\end{minipage}\hfill
\begin{minipage}[t]{.49\linewidth}\vspace{0pt}
We conduct a blinded study of WorldWeave and six world models on 20 selected scene--trajectory cases (140 videos). Participants score camera, object, and structure/texture consistency from 1 to 5. Each video receives three or four ratings, totaling 436 evaluations. Table~\ref{tab:user_study} reports video means averaged equally across cases (Appendix~\ref{app:user_study}). WorldWeave leads with 4.57, 4.73 and 4.52, compared with 3.92, 3.42 and 3.32 for Echo-WM. The gains over Echo-WM are 0.65, 1.31 and 1.20 points.
\end{minipage}\par

\section{Conclusion}
\label{sec:conclusion}
We presented WorldWeave, a world generation framework that decouples persistent structural memory from visual synthesis. Continual metric terrain expansion and agent-guided scene compilation extend an explicit world while preserving previously committed structure. Hierarchical planning connects user intent, terrain evidence and inherited interfaces; deterministic compilation, bounded revision and validated publication turn these decisions into reusable geometry. Read-only camera queries then provide depth conditions for video generation without writing synthesized observations back into world state. Across the evaluated methods, WorldWeave leads the structural-consistency and camera-compliance measures while retaining competitive visual quality. Human ratings on the 20 selected cases also favor WorldWeave in camera, object and structure/texture consistency. These results support maintaining a shared geometric world independently of individual video-generation windows, including observations with greater realized motion than most baselines.

\paragraph{Limitations and future work.}
Our current focus is static geometry, and depth provides guidance rather than a hard rendering constraint. Future work will address dynamic interactions, long-horizon appearance memory and adaptive spatial detail for larger worlds.

\medskip
\subsection*{AI Use Statement}
Generative AI tools assisted with drafting sections of the manuscript, organizing and polishing the writing, and literature retrieval and discovery, including identifying potentially relevant related work. The authors are responsible for checking AI-assisted text and references against the underlying sources, and for the accuracy, originality and integrity of the final manuscript.

\subsection*{Ethics Statement}
Our evaluation uses virtual environments and generated videos. Participants in the user study provide informed consent and evaluate anonymized method outputs. The terrain data and geometry assets are drawn from authored virtual environments, not presented as surveyed real-world geography. Third-party assets, pretrained models and services remain subject to their respective licenses and terms; redistribution requires the corresponding permissions. As with other visual generation systems, outputs could be used to create misleading depictions. Applications should clearly disclose synthetic content and preserve provenance, and should not treat generated worlds as verified representations of real locations.

\subsection*{Reproducibility Statement}
Section~\ref{sec:method} defines the world representation, incremental construction and read-only observation interfaces; Section~\ref{sec:experiments} specifies the evaluation setting and reports the principal comparisons. Appendices~\ref{app:terrain_details}--\ref{app:query_details} detail terrain encoding and joining, scene compilation, revision and geometric queries. Appendix~\ref{app:evaluation} defines the metrics and Appendix~\ref{app:ablation} specifies ablation protocols; Appendices~\ref{app:terrain_visual} and~\ref{app:seed_stability} provide matched terrain visualizations and per-task seed results. Appendix~\ref{app:user_study} specifies user-study selection and score aggregation.

\bibliography{references}
\bibliographystyle{plainnat}

\clearpage
\appendix
\setcounter{topnumber}{2}
\makeatletter\setlength{\@fptop}{0pt}\makeatother
\section{Supplementary Ablation Studies}
\label{app:ablations}
\subsection{Ablation Protocol}
\label{app:ablation}
The codec test compares five encodings on 64 matched tasks through one frozen VAE. Terrain input evaluation uses four neighbor configurations, 64 targets each and two protocols, matching encoding, inference steps and seeds. Canvas uses pooled-context training and multi-image models use context subsets; targets are in-domain authored RDR2 terrain. The default planning model is GPT-5.6-Sol (\texttt{gpt-5.6-sol}, low reasoning effort), compared with Qwen3.8-Max. For planning-model substitution, terrain, intent, asset catalog, compiler, revision budget and cameras are fixed. We compare relation realization, geometric validity and revision cost alongside asset layouts.

\subsection{Visual Comparison of Terrain Conditioning}
\label{app:terrain_visual}
\begin{figure}[p]
\centering
\setlength{\tabcolsep}{4pt}
\begin{tabular}{@{}cc@{}}
\textbf{Canvas} & \textbf{Multi-image}\\[3pt]
\multicolumn{2}{c}{(a) 1to1: one neighbor}\\
\includegraphics[width=.38\linewidth]{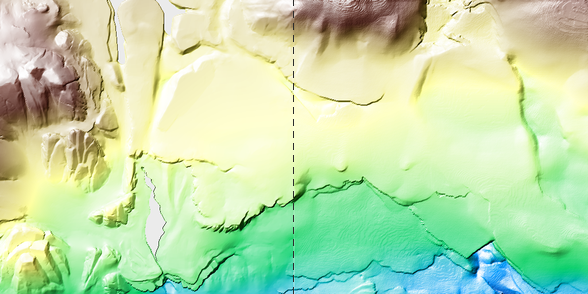}&
\includegraphics[width=.38\linewidth]{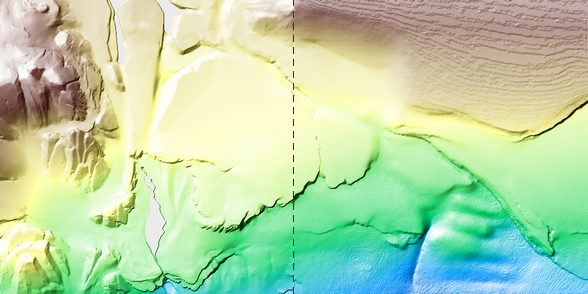}\\[3pt]
\multicolumn{2}{c}{(b) 2to1: opposite neighbors}\\
\includegraphics[width=.38\linewidth]{figures/terrain_2to1_opposite_canvas.png}&
\includegraphics[width=.38\linewidth]{figures/terrain_2to1_opposite_multi.png}\\[3pt]
\multicolumn{2}{c}{(c) L: adjacent neighbors}\\
\includegraphics[width=.38\linewidth]{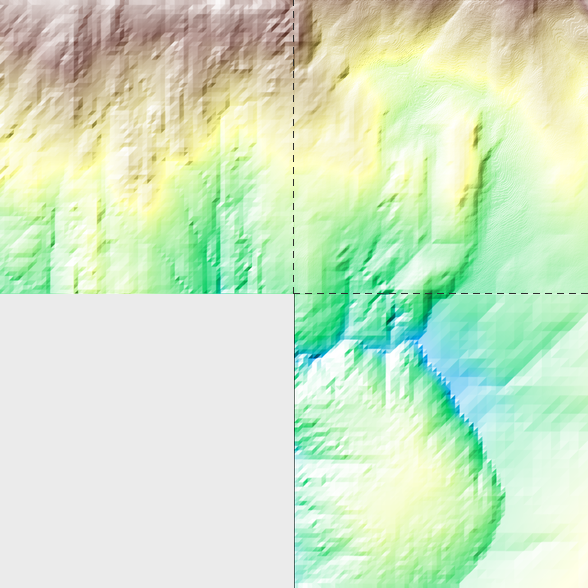}&
\includegraphics[width=.38\linewidth]{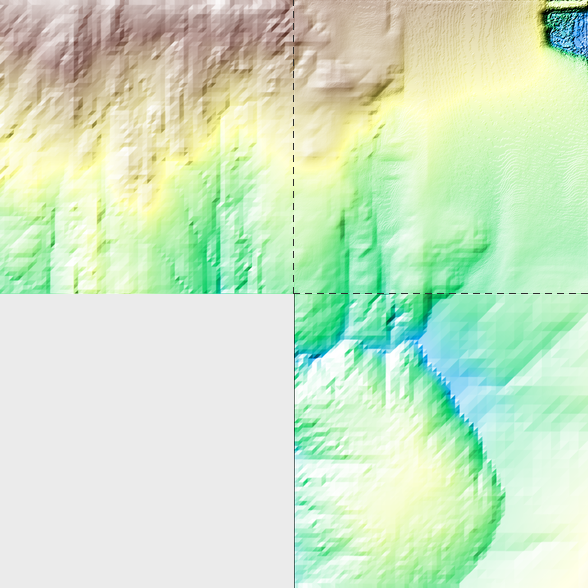}\\[3pt]
\multicolumn{2}{c}{(d) 3to1: three neighbors}\\
\includegraphics[width=.38\linewidth]{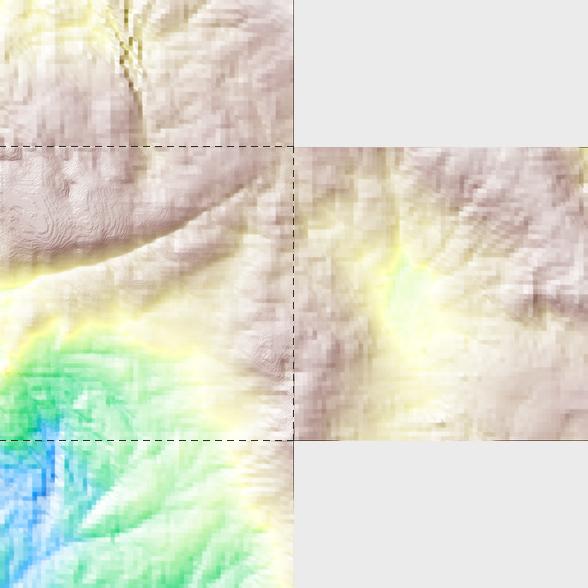}&
\includegraphics[width=.38\linewidth]{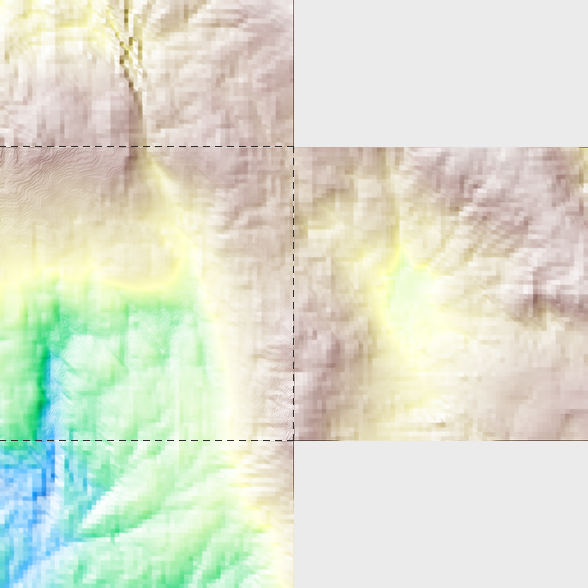}\\
\end{tabular}
\caption{Terrain completion with spatial canvas (left) and separate images (right). Each pair shares known terrain, elevation colors and illumination. Dashed lines mark context--target interfaces; gray regions are unavailable.}
\label{fig:terrain_input_visual}
\end{figure}

Figure~\ref{fig:terrain_input_visual} complements Table~\ref{tab:terrain_input_results}(b) with matched terrain completions. Spatial canvas conditioning maintains more coherent relief across the marked interfaces in these examples, supporting the boundary-continuity gains in the quantitative comparison.

\subsection{Planning-model Comparison}
\label{app:planner_comparison}
\begin{figure}[!htp]
\centering
\begin{minipage}[t]{.485\linewidth}
\centering
{\small (a) Alternative planner\par}
\includegraphics[width=\linewidth]{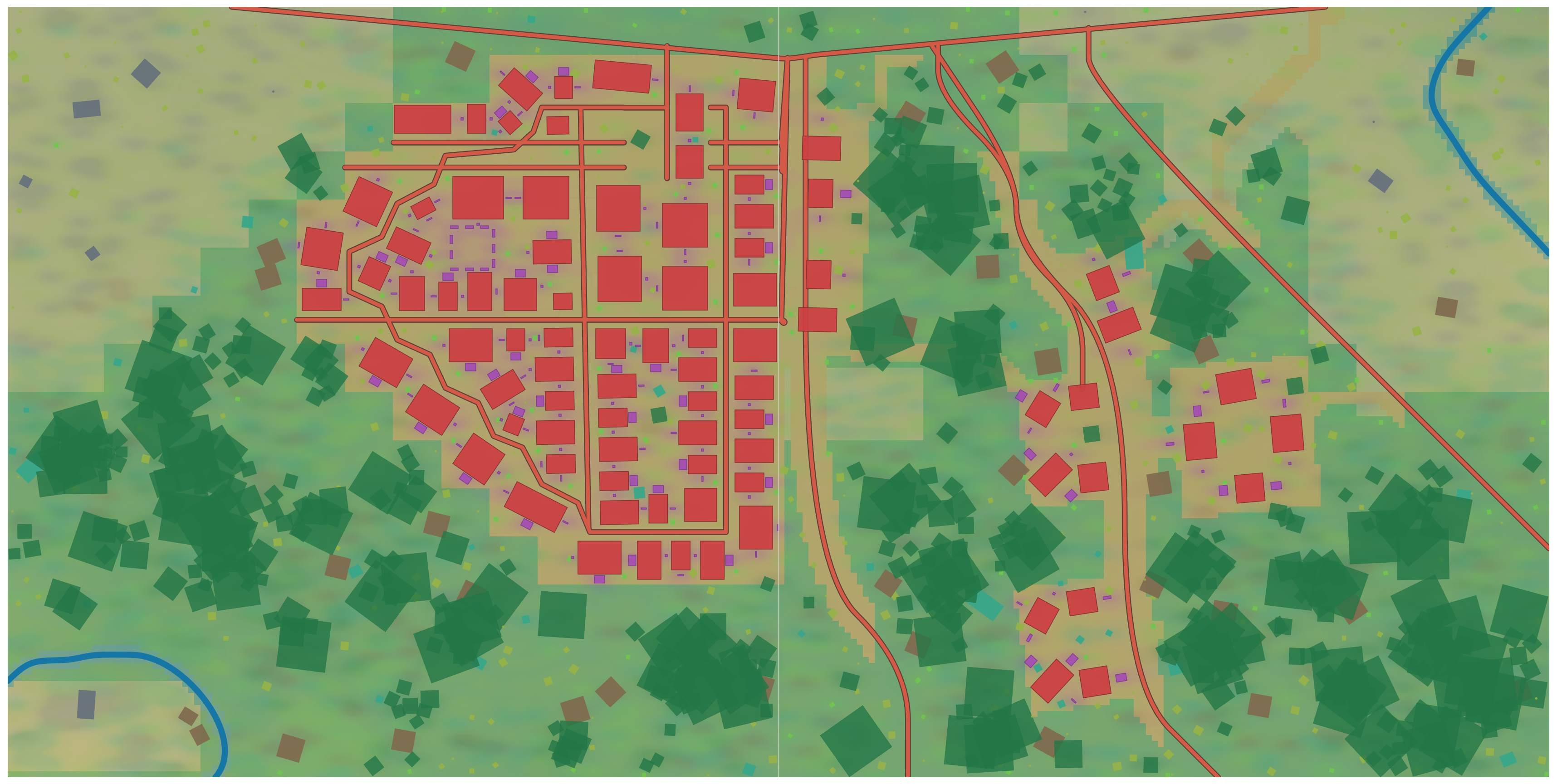}
\end{minipage}\hfill
\begin{minipage}[t]{.485\linewidth}
\centering
{\small (b) Default planner\par}
\includegraphics[width=\linewidth]{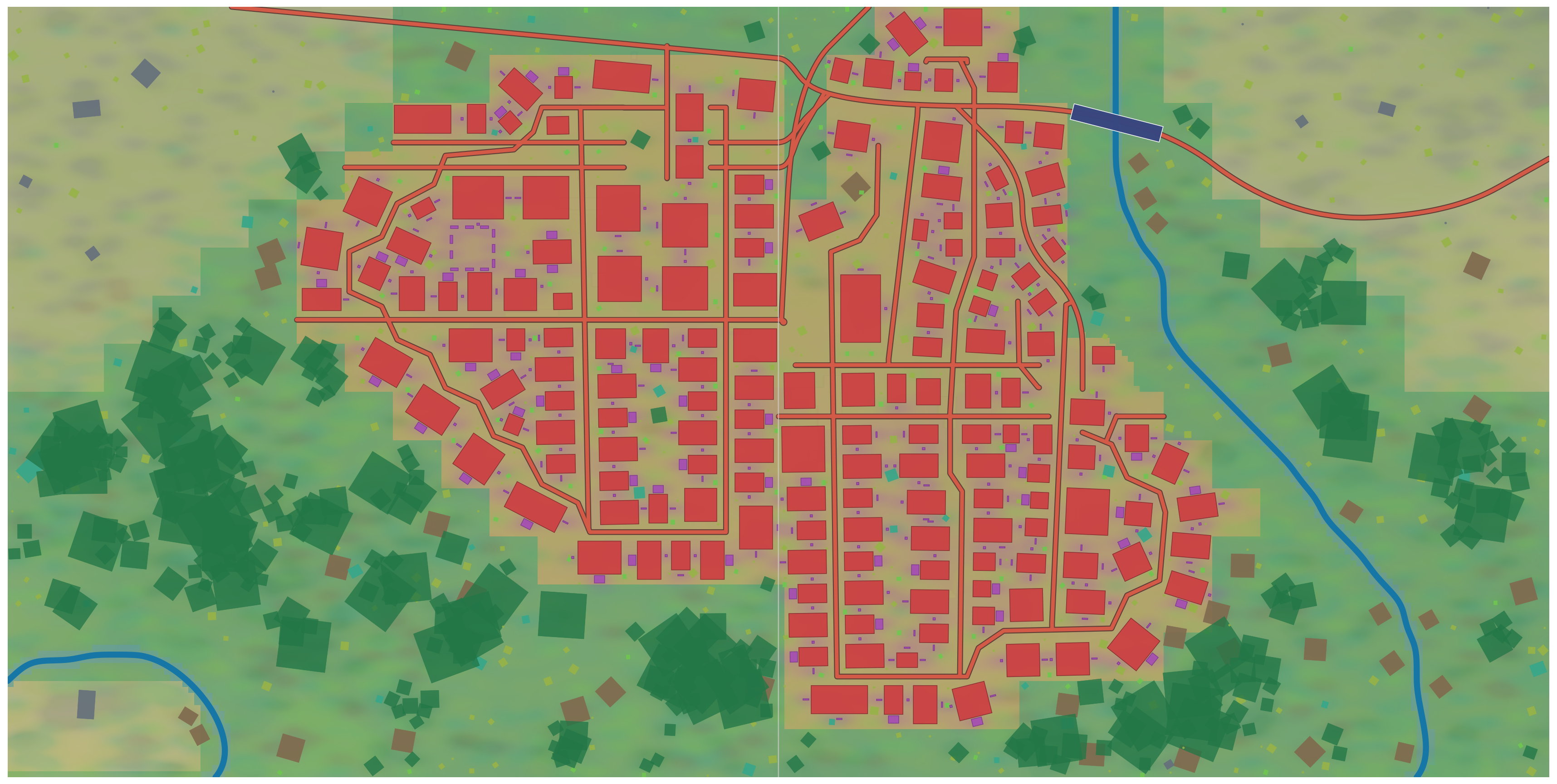}
\end{minipage}
\caption{Full planning-model comparison: (a) alternative and (b) default planner. Within each map, the shared existing region is on the left and the new region on the right. Terrain and drawing conventions are identical.}
\label{fig:agent_driver}
\end{figure}

Figure~\ref{fig:agent_driver} compares full layouts with identical terrain and an unchanged existing region on the left. The alternative planner favors wooded open space and dispersed settlements; the default planner produces a denser settlement and road network. Both retain the inherited terrain interface and organized connections.

\subsection{Random-seed Stability}
\label{app:seed_stability}
\subsubsection{Controlled Comparison}
We evaluate five fixed scene--trajectory pairs with seeds 0, 1, 2, 42 and 100. The world version, first frame, prompt, depth and non-seed generation settings are held fixed. Smoothness and MC-GeCo use the main comparison's frozen task-wise motion reference; MC-GeCo divides each video's co-visible residual by its relative motion before aggregation. Table~\ref{tab:seed_full} reports all 25 controlled outputs together with the five main-experiment references. Seed dispersion is computed from the 25 controlled outputs. 
\begin{table}[!htp]
\caption{Per-task seed comparison. Each scene contains one main-experiment reference and five controlled seed variants. All values are retained; $^\dagger$ marks the observed black-sky case.}
\label{tab:seed_full}
\centering\scriptsize
\setlength{\tabcolsep}{2pt}
\begin{tabular*}{\linewidth}{@{\extracolsep{\fill}}llrrrrrrrr@{}}
\toprule
Version & Seed & IQ$\uparrow$ & AQ$\uparrow$ & MN-MS$\uparrow$ & MC-GeCo$\downarrow$ & SMC$\uparrow$ & Cam$\uparrow$ & MC-MEt3R$\downarrow$ & MC-GeoCon$\downarrow$\\
\midrule
\multicolumn{10}{@{}l}{\textit{Forest}}\\
Reference & -- & 0.7927 & 0.5921 & 0.9853 & 0.0468 & 1.0000 & 97.62 & 0.1080 & 0.0768\\
Seed & 0 & 0.7943 & 0.5928 & 0.9834 & 0.0463 & 1.0000 & 97.62 & 0.1075 & 0.1002\\
Seed & 1 & 0.7917 & 0.5929 & 0.9825 & 0.0448 & 1.0000 & 97.62 & 0.1053 & 0.0957\\
Seed & 2 & 0.7916 & 0.5993 & 0.9828 & 0.0431 & 1.0000 & 97.62 & 0.1095 & 0.0902\\
Seed & 42 & 0.7896 & 0.5929 & 0.9847 & 0.0423 & 1.0000 & 97.62 & 0.1078 & 0.0860\\
Seed & 100 & 0.7930 & 0.6025 & 0.9862 & 0.0374 & 1.0000 & 95.24 & 0.1085 & 0.0655\\
\midrule
\multicolumn{10}{@{}l}{\textit{Rocky grassland}}\\
Reference & -- & 0.7930 & 0.6765 & 0.9842 & 0.0687 & 1.0000 & 97.62 & 0.1100 & 0.0660\\
Seed & 0 & 0.7859 & 0.6473 & 0.9850 & 0.0783 & 1.7500 & 97.62 & 0.1128 & 0.0871\\
Seed & 1 & 0.7874 & 0.6554 & 0.9840 & 0.0881 & 0.0000 & 40.48 & 0.1110 & 0.0806\\
Seed & 2 & 0.7873 & 0.6443 & 0.9862 & 0.0743 & 1.7500 & 97.62 & 0.1110 & 0.0814\\
Seed & 42 & 0.7918 & 0.6631 & 0.9853 & 0.0628 & 0.0000 & 40.48 & 0.1132 & 0.0779\\
Seed & 100 & 0.7854 & 0.6407 & 0.9859 & 0.0652 & 0.0000 & 32.14 & 0.1120 & 0.0805\\
\midrule
\multicolumn{10}{@{}l}{\textit{Town}}\\
Reference & -- & 0.7958 & 0.7683 & 0.9789 & 0.0513 & 2.0000 & 97.96 & 0.1357 & 0.1563\\
Seed & 0 & 0.7960 & 0.7641 & 0.9789 & 0.0496 & 1.5000 & 97.96 & 0.1378 & 0.1880\\
Seed & 1 & 0.7966 & 0.7795 & 0.9806 & 0.0526 & 0.0000 & 41.84 & 0.1335 & 0.1490\\
Seed & 2 & 0.7963 & 0.7716 & 0.9795 & 0.0524 & 1.0000 & 97.96 & 0.1355 & 0.1790\\
Seed & 42 & 0.7974 & 0.7683 & 0.9805 & 0.0530 & 1.5000 & 97.96 & 0.1348 & 0.1988\\
Seed & 100 & 0.7964 & 0.7714 & 0.9784 & 0.0502 & 1.5000 & 97.96 & 0.1363 & 0.1809\\
\midrule
\multicolumn{10}{@{}l}{\textit{Mountain}}\\
Reference & -- & 0.8030 & 0.7135 & 0.9809 & 0.0490 & 3.5000 & 97.96 & 0.1166 & 0.0985\\
Seed & 0 & 0.8044 & 0.7259 & 0.9810 & 0.0460 & 3.5000 & 97.96 & 0.1180 & 0.1274\\
Seed & 1 & 0.8027 & 0.7207 & 0.9811 & 0.0436 & 2.5000 & 97.96 & 0.1107 & 0.1027\\
Seed & 2 & 0.8021 & 0.7190 & 0.9807 & 0.0475 & 3.5000 & 97.96 & 0.1118 & 0.1300\\
Seed & 42 & 0.8028 & 0.7297 & 0.9819 & 0.0462 & 3.3750 & 97.96 & 0.1093 & 0.0891\\
Seed & 100$^\dagger$ & 0.7667 & 0.7153 & 0.9818 & 0.0431 & 3.5000 & 97.96 & 0.1576 & 0.0949\\
\midrule
\multicolumn{10}{@{}l}{\textit{Winter}}\\
Reference & -- & 0.7921 & 0.7029 & 0.9751 & 0.0525 & 0.0000 & 34.69 & 0.1764 & 0.1709\\
Seed & 0 & 0.7922 & 0.6954 & 0.9772 & 0.0403 & 1.0000 & 97.96 & 0.1669 & 0.1817\\
Seed & 1 & 0.7899 & 0.6930 & 0.9759 & 0.0451 & 0.0000 & 34.69 & 0.1634 & 0.1607\\
Seed & 2 & 0.7886 & 0.6785 & 0.9777 & 0.0384 & 0.0000 & 34.69 & 0.1638 & 0.1661\\
Seed & 42 & 0.7943 & 0.6975 & 0.9777 & 0.0388 & 0.0000 & 34.69 & 0.1612 & 0.1583\\
Seed & 100 & 0.7914 & 0.6953 & 0.9783 & 0.0399 & 0.0000 & 34.69 & 0.1638 & 0.1555\\
\bottomrule
\end{tabular*}
\end{table}
\subsubsection{Metric Variation and Visual Outcomes}
Table~\ref{tab:seed_summary} first averages the same five tasks for each seed. Its standard deviation and coefficient of variation are computed across those five seed-wise means, using the population standard deviation. This separates aggregate seed sensitivity from differences between scenes.
\begin{table}[!htp]
\caption{Seed-wise means over five tasks and their across-seed dispersion. The five main-experiment references are excluded. SD is the population standard deviation and CV is SD divided by the mean, expressed as a percentage.}
\label{tab:seed_summary}
\centering\scriptsize
\setlength{\tabcolsep}{2pt}
\begin{tabular*}{\linewidth}{@{\extracolsep{\fill}}lrrrrrrrr@{}}
\toprule
Seed & IQ & AQ & MN-MS & MC-GeCo & SMC & Cam & MC-MEt3R & MC-GeoCon\\
\midrule
0 & 0.7945 & 0.6851 & 0.9811 & 0.0521 & 1.7500 & 97.82 & 0.1286 & 0.1369\\
1 & 0.7937 & 0.6883 & 0.9808 & 0.0548 & 0.7000 & 62.52 & 0.1248 & 0.1177\\
2 & 0.7932 & 0.6825 & 0.9814 & 0.0512 & 1.4500 & 85.17 & 0.1263 & 0.1293\\
42 & 0.7952 & 0.6903 & 0.9820 & 0.0486 & 1.1750 & 73.74 & 0.1253 & 0.1220\\
100 & 0.7866 & 0.6850 & 0.9821 & 0.0472 & 1.2000 & 71.60 & 0.1356 & 0.1155\\
\midrule
Mean & 0.7926 & 0.6863 & 0.9815 & 0.0508 & 1.2550 & 78.17 & 0.1281 & 0.1243\\
SD & 0.0031 & 0.0027 & 0.0005 & 0.0027 & 0.3466 & 12.19 & 0.0040 & 0.0079\\
CV (\%) & 0.39 & 0.40 & 0.05 & 5.29 & 27.61 & 15.60 & 3.11 & 6.33\\
\bottomrule
\end{tabular*}
\end{table}

\paragraph{Computational measures.}
The aggregate IQ, AQ and MN-MS vary narrowly across seeds, with CVs of 0.39\%, 0.40\% and 0.05\%. The geometric scores have larger but still comparatively moderate CVs: 5.29\% for MC-GeCo, 3.11\% for MC-MEt3R and 6.33\% for MC-GeoCon. These results support stability of the aggregate appearance and geometry measures under the tested noise changes. Per-task results remain important: averaging can reduce dispersion and does not imply that every individual video is equally stable.

\paragraph{Occasional sky corruption.}
Visual inspection identifies a black-sky interval in one of the 25 controlled outputs, the mountain scene with seed 100. Its IQ is 0.7667, compared with 0.8021--0.8044 for the other four seeds, and MC-MEt3R rises to 0.1576 from 0.1093--0.1180. The outcome is consistent with the base video generator occasionally transferring dark depth-conditioning regions into sky appearance, although the seed comparison alone does not isolate that cause. The affected video remains in all statistics. This observation identifies an appearance-synthesis failure under unchanged geometry; one occurrence in this small study does not establish its deployment frequency.

\paragraph{Event-based evaluation.}
SMC and Cam have CVs of 27.61\% and 15.60\%, respectively, and vary more than the computational measures. Their discrete disappearance, return and motion-clause decisions can amplify small changes in generated evidence; the visual judgments additionally involve a large-model evaluator. These scores therefore reflect event realization and evaluation sensitivity as well as visual consistency. The present experiment varies generation seeds, not repeated judgments of an identical video, so it does not attribute all dispersion to evaluator randomness. We report their full ranges alongside the more stable computational measures.

\section{Evaluation Metrics}
\label{app:evaluation}

\subsection{Metric Definitions}
\label{app:metrics}
\textbf{Visual quality.} IQ averages MUSIQ frame-quality estimates, AQ averages CLIP-based aesthetic predictions, and MS measures agreement with AMT-interpolated frames, following VBench~\citep{vbench2024}. All measures evaluate generated RGB under the same public task description.

\textbf{Structural Memory Consistency (SMC).} SMC requires dedicated camera trajectories that induce a disappearance--return event for a designated target. To enable this evaluation, we randomly select 40\% of all samples and adapt their sampling trajectories and camera motions accordingly; SMC is computed on this subset. A method-blind Qwen3.5-27B evaluator samples each video at 1\,Hz, identifies the named target and its neighbors, and selects the earliest usable reference. It verifies complete target absence, then compares the reference with the earliest and latest usable return views. Each pair receives anchored 1--5 grades for identity, structure, neighbor relations and texture: 5 denotes preservation, 3 a clear local change, and 1 replacement or unrecognizable structure. Legitimate viewpoint and lighting changes are allowed. With grades $g_{jd}$ for return view $j$ and criterion $d$, the score is
\begin{equation}
 s_j=\operatorname{Cap}\!\left(\frac{\sum_{d\in D_j}w_d g_{jd}}{\sum_{d\in D_j}w_d}\right),\qquad
 \mathrm{SMC}=\frac{1}{|\mathcal{R}|}\sum_{j\in\mathcal{R}}s_j,\qquad
 w=(0.30,0.30,0.25,0.15).
 \label{eq:smc_score}
\end{equation}
Here $D_j$ contains graded criteria and $\mathcal{R}$ the return views. Identity, structure and neighbor evidence are required; an unobservable texture grade is omitted. Identity grade 1 fixes the pair score at 1; grade 2 caps it at 2. Structure at most 2 caps it at 2.5, and neighbor relations at most 2 cap it at 3. A designated memory task scores zero when its required disappearance--return event is absent.

\textbf{Camera Compliance (Cam).} The public prompt is decomposed into motion and visibility clauses. Translation and turning are evaluated from estimated camera poses at 1\,Hz; target framing, absence and return use visual evidence. An event clause receives 1 when observed and 0 otherwise; a sustained clause receives the fraction of intervals satisfying it. Clause degrees $d_c$ are combined as
\begin{equation}
 \mathrm{Cam}=\frac{100\,o}{|\mathcal{C}|}\sum_{c\in\mathcal{C}}d_c,\qquad
 o=1\ \text{for correct event order},\quad o=\tfrac12\ \text{otherwise}.
 \label{eq:camera_score}
\end{equation}
Thus Cam rewards executing the requested observation sequence, while SMC evaluates what is preserved when the target returns.

\textbf{Geometric consistency.} GeCo combines motion and depth residuals~\citep{geco2025}; MEt3R compares reprojected image features~\citep{met3r2025}; GeoCon-Bench measures correspondence inlier ratio and geometric fitting error~\citep{geocon2026}. We use a correspondence-based local reproduction for GeoCon and apply the corrections below.

\subsection{Motion-Calibrated Consistency and Smoothness}
\label{app:corrections}
\paragraph{Realized motion.}
Motion amplitude is the evaluator's mean optical-flow norm, reflecting scene depth, camera rotation and translation, and scene motion. Across 120 paired tasks, WorldWeave exceeds the six baselines listed in the main text on 74--93 tasks each.

\begin{figure}[!htp]
\centering
\setlength{\tabcolsep}{1pt}
\footnotesize
\begin{tabular}{@{}cccc@{}}
\multicolumn{4}{l}{(a) Texture degradation: \texttt{dev-winter-a}}\\
\multicolumn{2}{c}{WorldWeave} & \multicolumn{2}{c}{AlayaWorld}\\
\includegraphics[width=.243\linewidth]{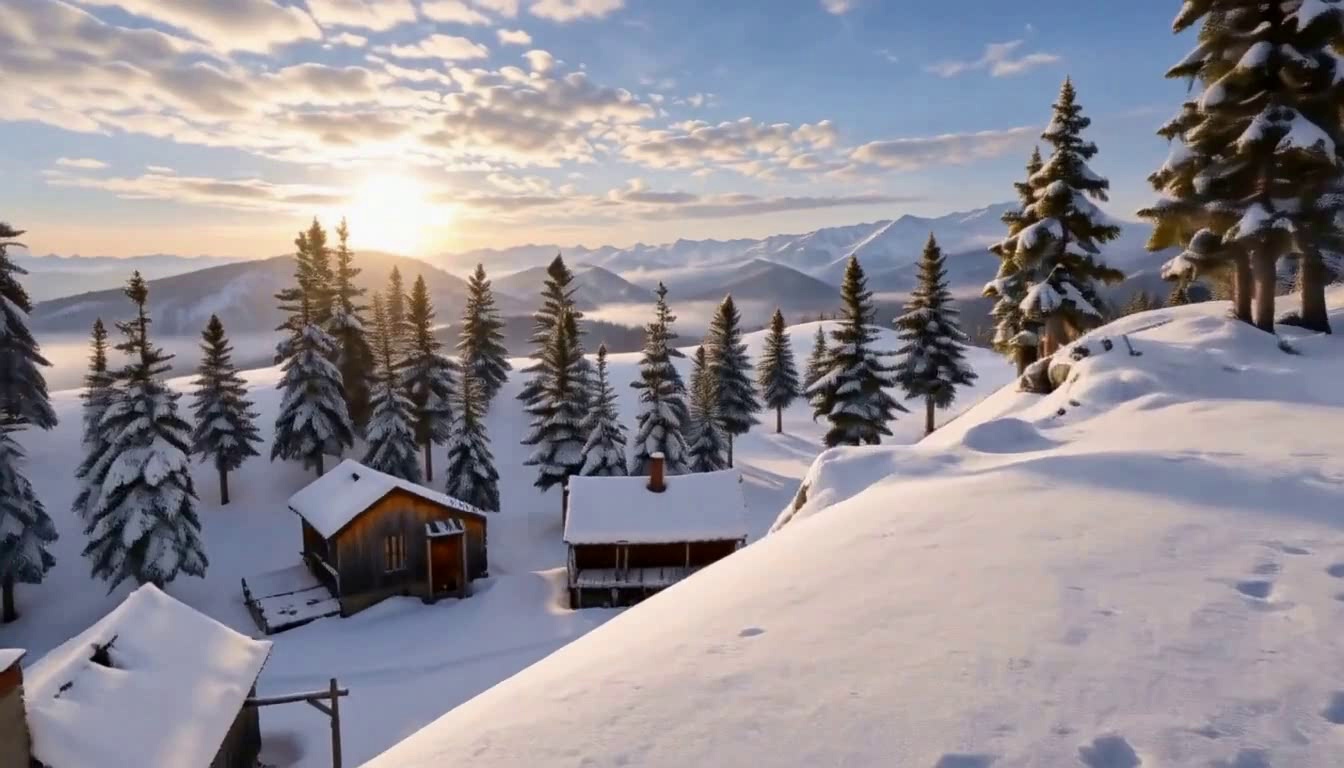}&
\includegraphics[width=.243\linewidth]{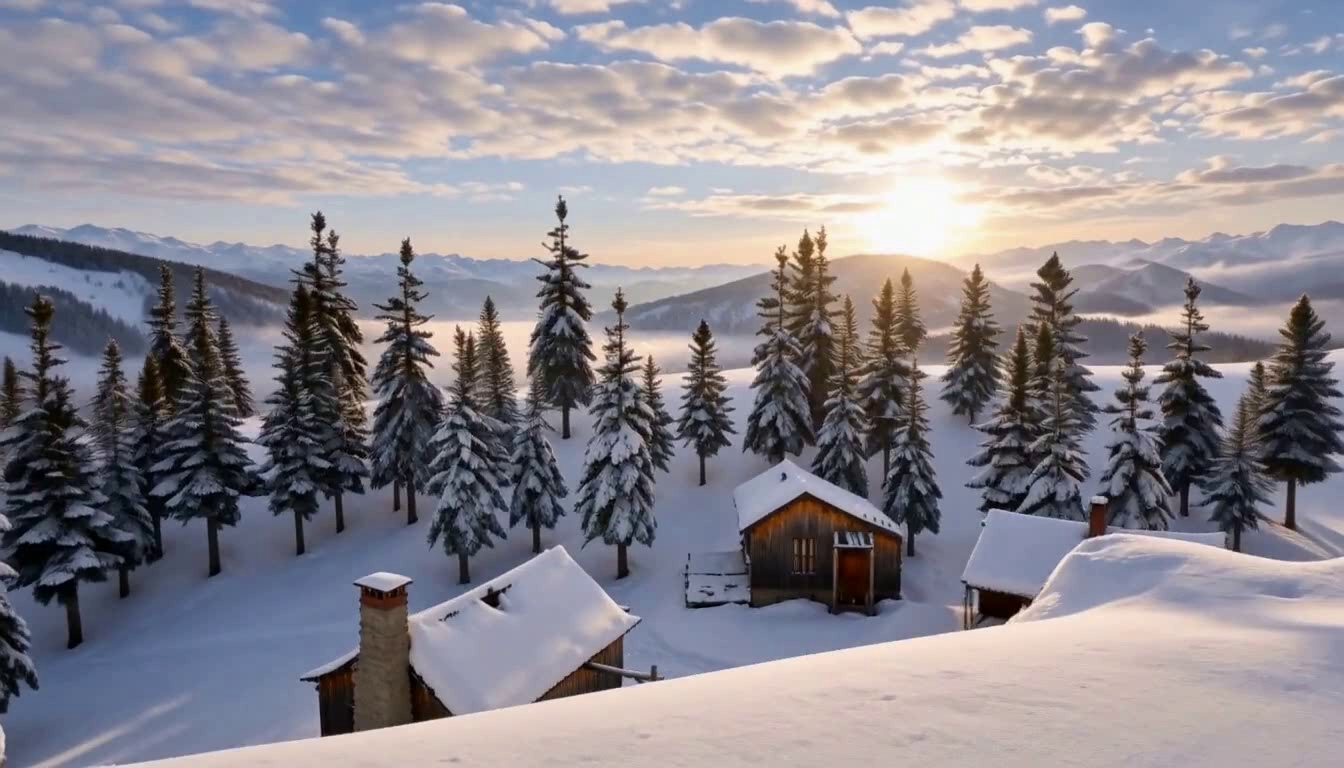}&
\includegraphics[width=.243\linewidth]{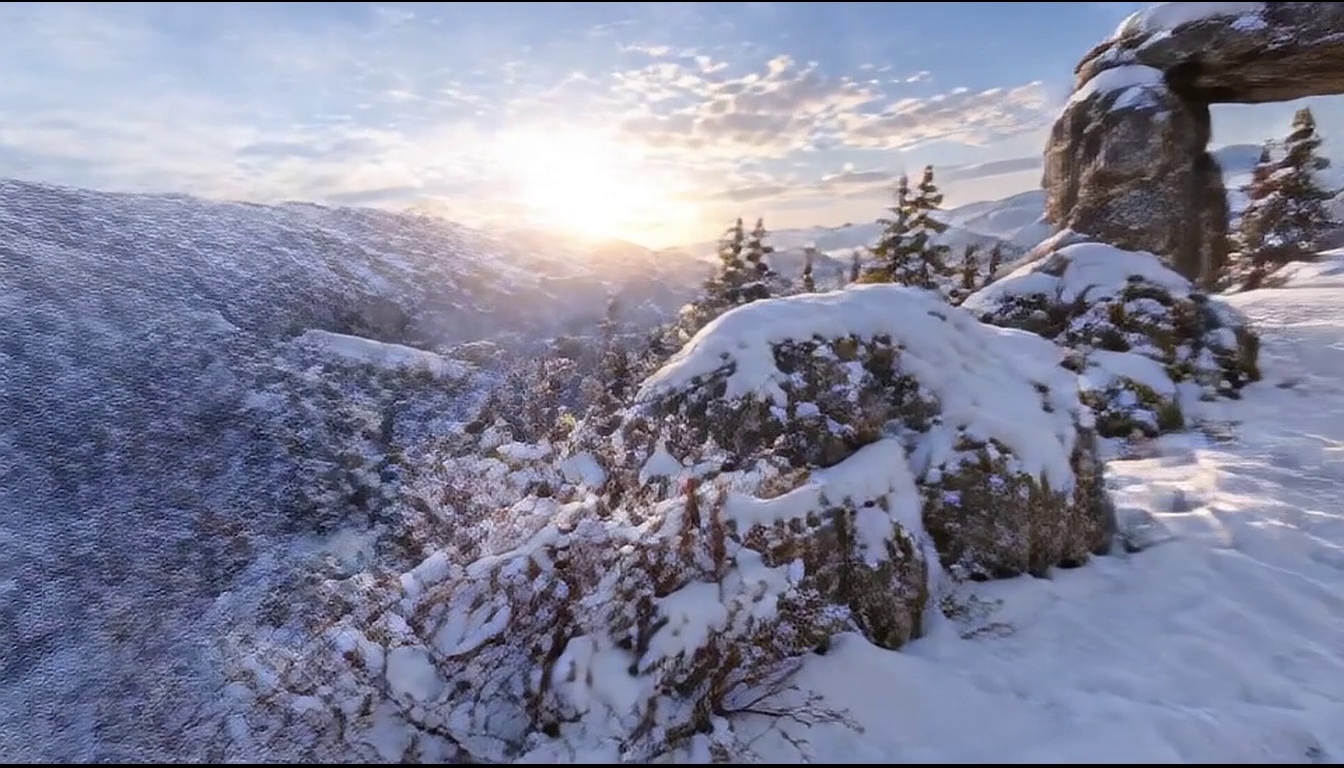}&
\includegraphics[width=.243\linewidth]{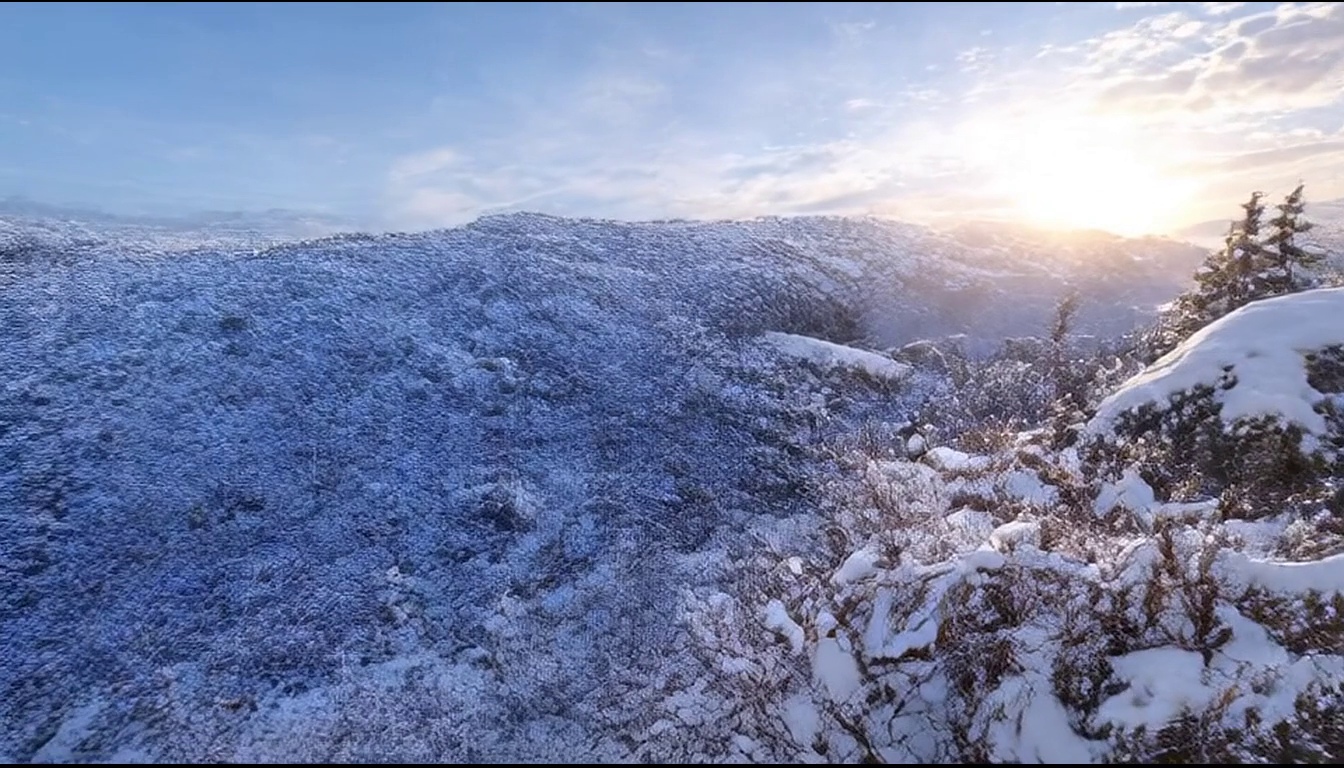}\\
\multicolumn{4}{l}{(b) Limited camera motion: \texttt{winter-04}}\\
\multicolumn{2}{c}{WorldWeave} & \multicolumn{2}{c}{Wan3.0}\\
\includegraphics[width=.243\linewidth]{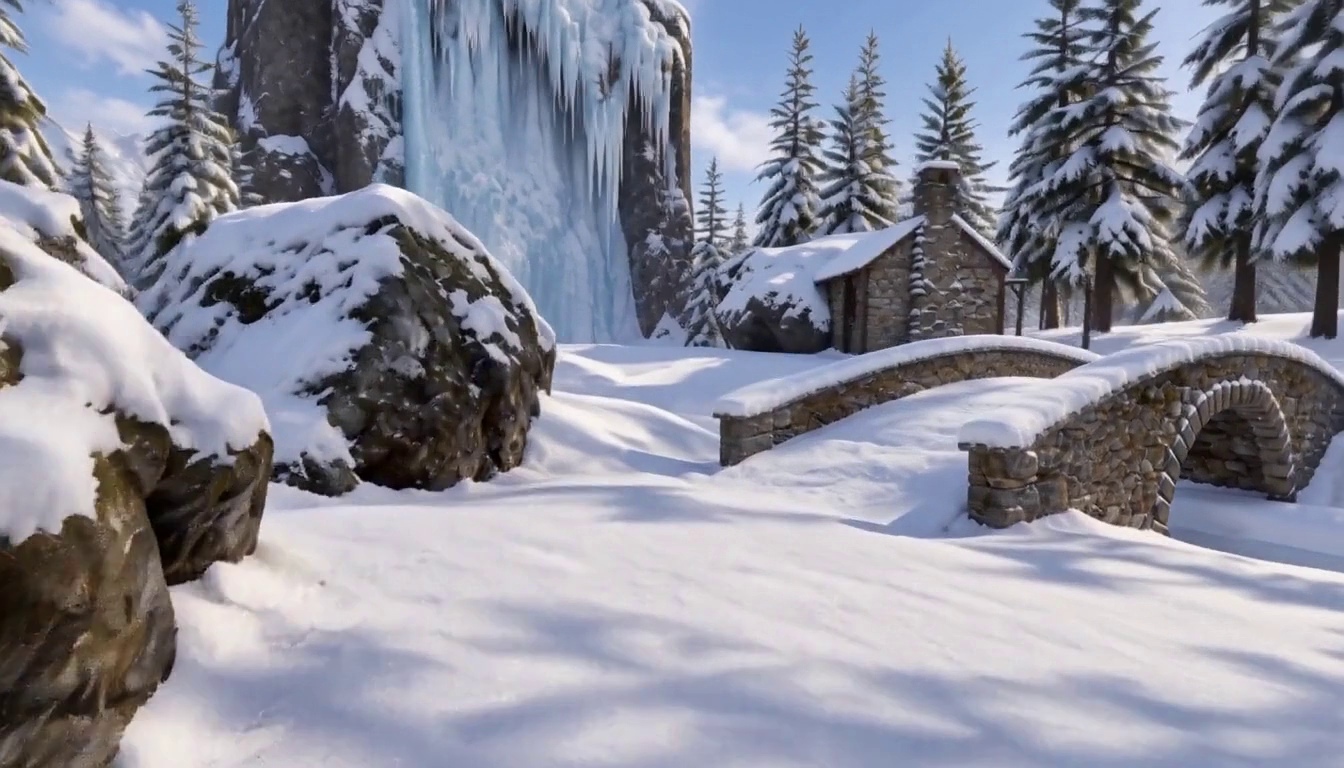}&
\includegraphics[width=.243\linewidth]{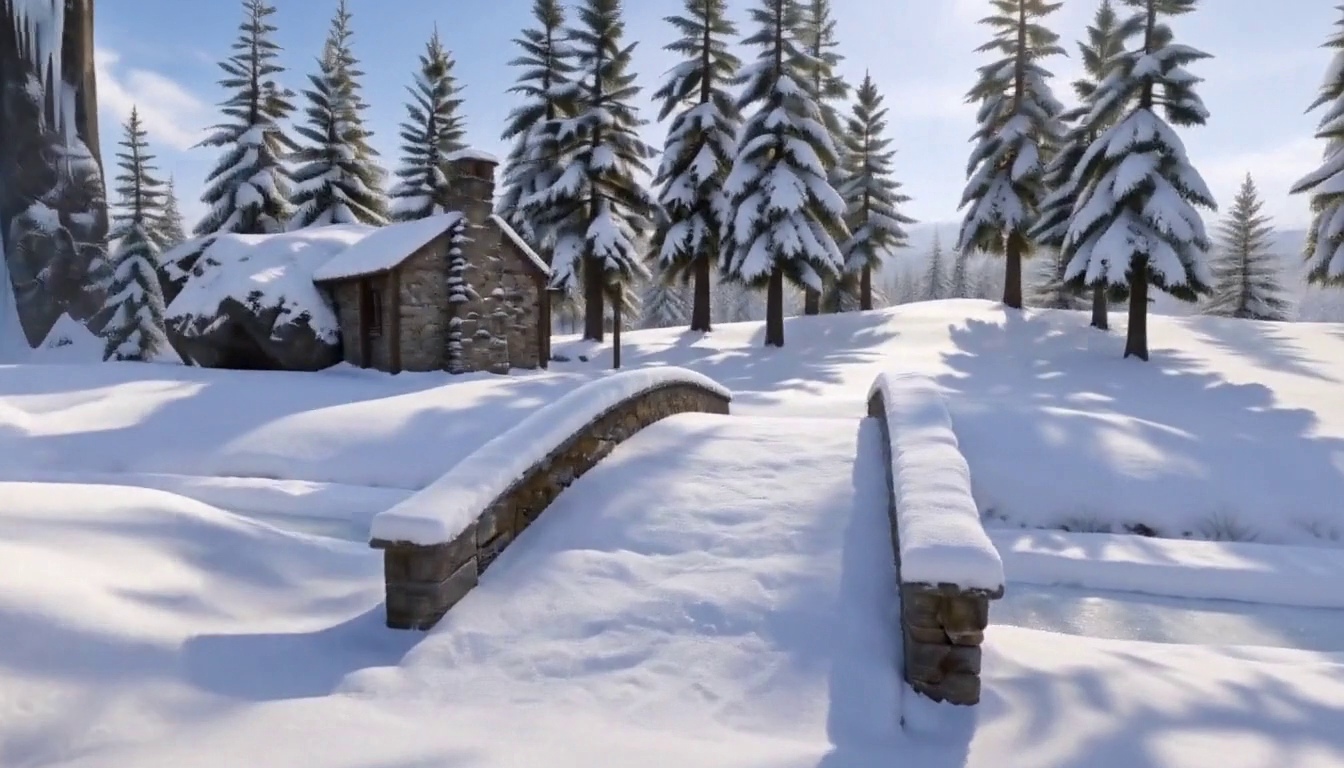}&
\includegraphics[width=.243\linewidth]{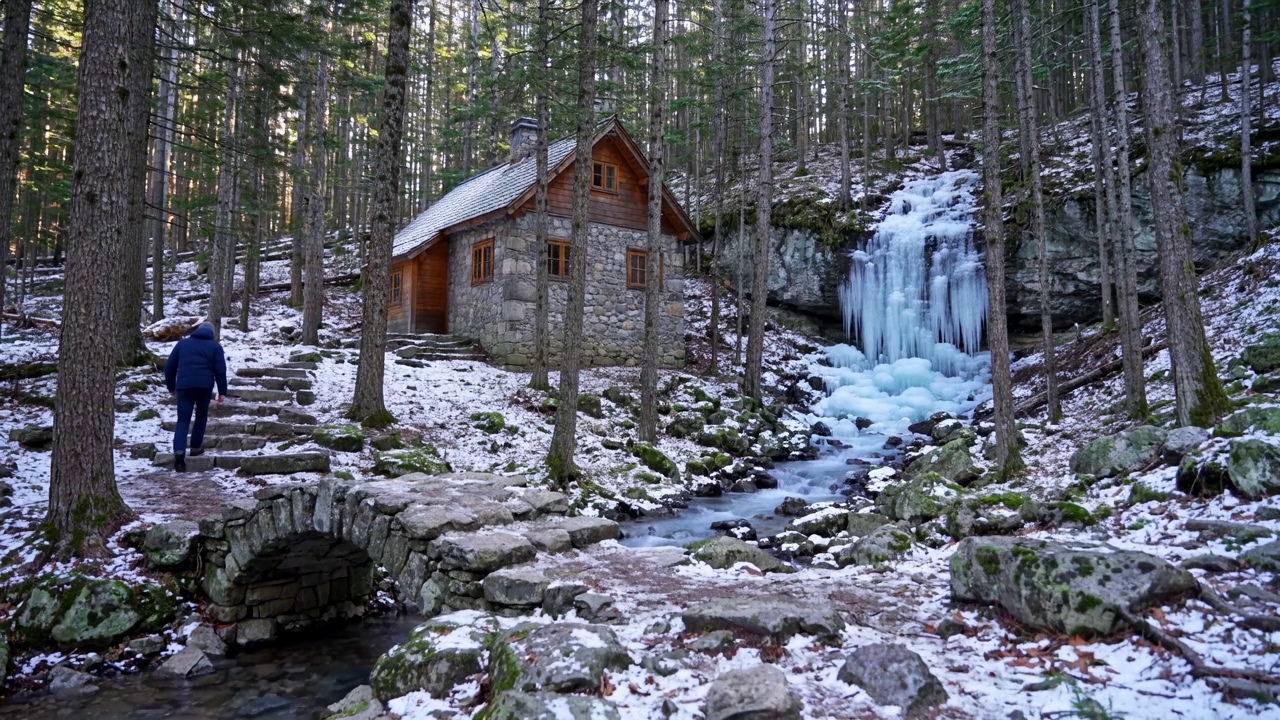}&
\includegraphics[width=.243\linewidth]{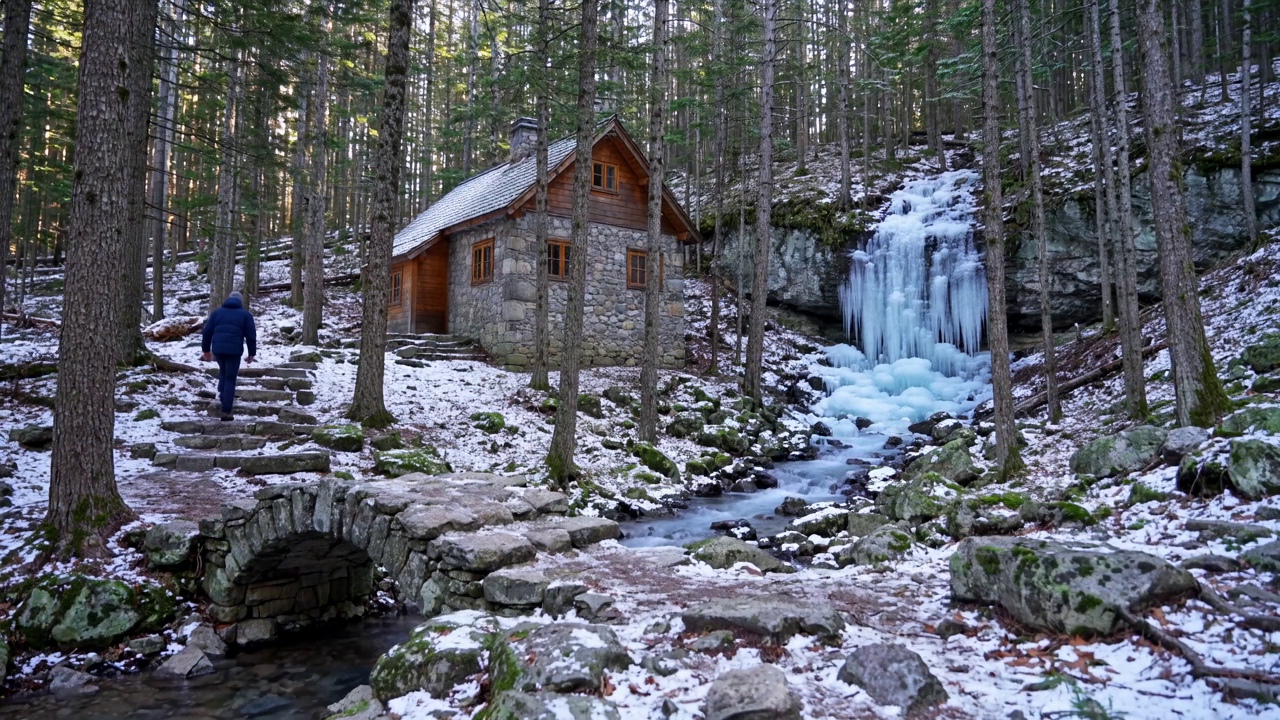}\\
\end{tabular}
\caption{Motion and consistency are complementary. AlayaWorld's low raw MEt3R error coexists with distorted snow texture; Wan3.0's low raw GeCo error accompanies little camera movement. Times are shown; scores refer to whole videos.}
\label{fig:metric_diagnostics}
\end{figure}

The task requires both scene preservation and camera movement. Absolute disagreement can decrease when a model barely changes its view, even though it misses the requested exploration. Figure~\ref{fig:metric_diagnostics} shows this mismatch: favorable raw scores coexist with distorted texture or nearly stationary views. We therefore measure geometric disagreement relative to the motion actually realized and evaluate camera compliance separately.

\textbf{Relative residuals on common visibility.} Let $u$ be observed angular flow, $u_r$ the flow induced by estimated camera motion and depth, and $z_w,z_t$ warped and target depths. On pixels that project inside the target and pass depth agreement, the co-visible GeCo residual uses
\begin{equation}
 e_m=\frac{\|u-u_r\|_2}{\|u\|_2+\|u_r\|_2+\epsilon},\qquad
 e_z=\frac{|z_w-z_t|}{|z_w|+|z_t|+\epsilon}.
 \label{eq:relative_geco}
\end{equation}
The flow denominator expresses deformation relative to observed and predicted displacement; the visibility mask prevents newly exposed content from being treated as a correspondence failure. Valid motion and depth cues are averaged equally, retaining GeCo's single-cue fallback.

\textbf{Task-relative motion normalization.} For GeCo, $e$ is the co-visible relative residual above; for MEt3R and GeoCon, it is the original feature error or combined geometric error. Each video $v$ is normalized by realized motion $m(v)$ relative to the fixed task-wise reference $m_{\mathrm{ref}}$, then scores are averaged:
\begin{equation}
\begin{aligned}
 r_m(v)&=\frac{m(v)}{m_{\mathrm{ref}}},&
 e_{\mathrm{MC}}(v)&=\frac{e(v)}{r_m(v)},\\
 \overline e_{\mathrm{MC}}&=\frac{1}{N}\sum_{i=1}^{N}e_{\mathrm{MC}}(v_i),&
 e_{\mathrm{GeoCon}}&=\sqrt{\max(1-\mathrm{IR},\epsilon)\max(\mathrm{GE},\epsilon)}.
\end{aligned}
\label{eq:motion_correction}
\end{equation}
This reports error per relative amount of exploration: small raw error receives less credit when accompanied by little movement. The reference is the task-wise median from the frozen comparison set, shared across methods and held fixed when adding the Base or seed variants. Aggregation uses the mean of video-wise ratios, not the ratio of aggregate errors and motion. Non-positive motion and invalid correspondences are excluded. GeoCon remains supplementary because the local reproduction is insensitive to the tested local warp. The geometric scores and Cam jointly assess preservation under completed camera motion.

\paragraph{Motion-normalized smoothness.}
We apply the same task-relative motion calibration to VBench's smoothness error:
\begin{equation}
\mathrm{MN\text{-}MS}(v)=1-\frac{1-\mathrm{MS}(v)}{r_m(v)} .
\label{eq:mn_ms}
\end{equation}
Higher is better. Scores are computed per video before aggregation and interpreted jointly with geometric consistency and camera compliance.

\section{User-study Protocol}
\label{app:user_study}
\paragraph{Cases and presentation.}
The study contains 20 matched scene--trajectory cases with seven methods per case. Cases were selected using automatic metrics and visual screening before human rating; the reported means characterize this curated evaluation set. Participants view anonymized videos under shared task instructions; method names and automatic scores are hidden. Each account is assigned ten distinct videos, with presentation order shuffled and allocation prioritizing rating coverage. Participation is voluntary and requires informed consent.

\paragraph{Rating criteria.}
Participants assign integer scores from 1 to 5, with higher scores indicating better consistency. Camera consistency evaluates the requested movement, turns and observation order. Object consistency evaluates preservation of object identity, shape and number. Structure/texture consistency evaluates stable scene layout and coherent surface details. The same three questions accompany every video.

\paragraph{Aggregation.}
All 140 videos meet the minimum of three ratings: 124 have three and 16 have four, yielding 436 video evaluations and 1,308 criterion scores. We first average ratings per video and criterion, then average the 20 cases with equal weight. Let $s_{cmdr}$ be rating $r$ for case $c$, method $m$ and criterion $d$, with $n_{cm}$ ratings per video. The method-level mean is
\begin{equation}
\bar s_{cmd}=\frac{1}{n_{cm}}\sum_{r=1}^{n_{cm}}s_{cmdr},\qquad
 \mu_d(m)=\frac{1}{|\mathcal C|}\sum_{c\in\mathcal C}\bar s_{cmd}.
\label{eq:user_study_mean}
\end{equation}
Here $\mathcal C$ contains all 20 cases. Each case has equal weight, so videos with four ratings do not contribute more than those with three. Scores remain on the original 1--5 scale; higher values indicate stronger perceived consistency. WorldWeave attains the unique highest object mean in 19 of 20 cases, and attains or shares the highest camera and structure/texture means in 16 and 18 cases.

\section{Metric Terrain Implementation}
\label{app:terrain_details}

\subsection{Context Reference and Coordinate Convention}

Let $\Delta$ be the metric sample spacing and $o_q$ the target origin. Sample centers are $x_{ij}=o_q+\Delta(i+\tfrac12,j+\tfrac12)$. Rotation into a canonical neighbor arrangement acts jointly on elevation, validity and side labels; outputs are mapped back before joining. Let $\mathcal{B}_{\mathrm{ctx}}$ collect samples from inward-facing context bands and $X_a=(1,x_a,y_a)$. The reference plane is initialized by least squares and refined by reweighted fits:

\begin{equation}
\begin{aligned}
 e_a^{(l)}&=X_a\theta^{(l)}-h_a,\quad
 \sigma_l=\max\{\sigma_{\min},1.4826\,\operatorname{median}_a|e_a^{(l)}|\},\\
 \omega_a^{(l)}&=\min\{1,1.5\sigma_l/\max(|e_a^{(l)}|,\epsilon)\},\\
 \theta^{(l+1)}&=\arg\min_\theta
 \sum_{a\in\mathcal{B}_{\mathrm{ctx}}}(\omega_a^{(l)})^2(X_a\theta-h_a)^2 .
\end{aligned}
\label{eq:app_reference_fit}
\end{equation}

The squared weights reflect weighting both the design matrix and observations. Multiple neighbors share one fit. With a single neighbor, the normal trend is tapered away from the interface to prevent indefinite linear extrapolation. Writing the plane coefficients as $(a,b,c)$ in normal--tangential coordinates $(d,y)$ gives

\begin{equation}
b_q(d,y)=a+cy+b\,\psi_L(d),\qquad
\psi_L(d)=
\begin{cases}
d-d^2/(2L),&0\le d<L,\\
L/2,&d\ge L.
\end{cases}
\label{eq:app_reference_taper}
\end{equation}

The normal derivative therefore decreases continuously to zero. Multi-neighbor configurations use the joint plane without this single-edge taper. Context bands are 64 samples wide; the fit uses four reweighting passes, $\sigma_{\min}=1\,\mathrm{m}$ and $L=64\,\mathrm{m}$. Missing elevations are filled from the nearest valid sample for numerical processing, while their original validity remains excluded from supervision. Target elevations never enter the context fit.

\subsection{Redundant Encoding and Robust Metric Decoding}

The channel map $f_j(r)=\tfrac12+\tfrac12\tanh(r/s_j)$ defines a one-dimensional curve in RGB space. Spatial repetition adds image-space redundancy, and block averaging restores the metric grid after VAE decoding:

\begin{equation}
(U_\rho c)_{\rho i+a,\rho j+b}=c_{ij},\qquad
(A_\rho\widetilde c)_{ij}=\rho^{-2}
\sum_{a,b=0}^{\rho-1}\widetilde c_{\rho i+a,\rho j+b},\qquad \rho=2.
\label{eq:app_repeat}
\end{equation}

For an averaged decoded sample $c_j$, clip channel endpoints before inversion. Define $y_j=2c_j-1$ and use inverse-sensitivity weighting to initialize the residual:

\begin{equation}
r_j=s_j\operatorname{atanh}(y_j),\quad
G_j=\frac{2s_j}{\max(1-y_j^2,\epsilon)},\quad
r^{(0)}=\frac{\sum_jG_j^{-2}r_j}{\sum_jG_j^{-2}}.
\label{eq:app_decode_init}
\end{equation}

Saturated channels have larger inverse gain and receive less weight. The channels need not agree after image decoding, so a fixed robust projection brings the estimate back toward their common encoding curve. With $J_j(r)=f'_j(r)$ and $\eta_j^{(l)}=\min\{1,\kappa/\max(|f_j(r^{(l)})-c_j|,\epsilon)\}$, the scalar update is

\begin{equation}
r^{(l+1)}=r^{(l)}-
\frac{\sum_j\eta_j^{(l)}J_j(r^{(l)})[f_j(r^{(l)})-c_j]}
{\max\{\sum_j\eta_j^{(l)}J_j(r^{(l)})^2,\epsilon\}},
\qquad \widehat h_q=b_q+r^{(L_d)}.
\label{eq:app_decode_project}
\end{equation}

We use $L_d=5$ and $\kappa=2/255$, clipping channel values to $[10^{-5},1-10^{-5}]$. The projection is determined by decoded channels and the analytic codec.

\subsection{Conditioning, Supervision and Boundary Acceptance}

Let $K_q,T_q,V_q$ denote known-context, target and valid masks on the spatial canvas. The condition retains encoded context and replaces target content with a neutral value. A separate region image distinguishes target, validity and unavailable locations:

\begin{equation}
C_{\mathrm{cond}}(x)=
\begin{cases}
f(H(x)-B(x)),&x\in K_q,\\
(1/2,1/2,1/2),&x\in T_q,\\
(0,0,1),&x\notin K_q\cup T_q,
\end{cases}
\quad
R_q=(T_qV_q,V_q,1-K_q-T_q).
\label{eq:app_canvas}
\end{equation}

Canvas slots preserve neighbor direction; half-neighbor bands are used where the canonical layout places opposite neighbors around the full target. Only target-valid locations contribute to the normalized loss in the main text. Distances for edge weights are evaluated in target-local metric coordinates, so image repetition and the loss-grid resolution do not change the physical correction-band width.

Joining is evaluated on the shared interface, not by equating distinct sample centers on opposite sides. Let $\gamma_e$ be interface height and derivative extraction with a common normal. Its residual is

\begin{equation}
E_{\partial q}(h)=
\left\{\gamma_e(h)-(g_e,v_e)\right\}_{e\in\mathcal E_q},
\quad
\gamma_e(h)=(h|_{\Gamma_e},\partial_{n_e}h|_{\Gamma_e}).
\label{eq:app_boundary_residual}
\end{equation}

The connector inherits height and derivatives at the boundary of each preserved old surface. Corrections vanish with their normal derivatives at their inner support boundaries, while corner reconciliation acts on new-owned nodes. Continuity and modification magnitude are measured separately; all targets remain in the joining comparison, including those requiring large corrections. 

\subsection{Ownership-aware Continuous Joining}
\label{app:joining_details}
Committed geometry is defined over the cells between existing sample centers. The gap between neighboring center grids and uncovered corner cells belong to the new connector surface. Let $J=(h,h_x,h_y,h_{xy})$ denote nodal height and derivatives. Bicubic Hermite connector cells inherit $J$ on their old-facing edges; new-owned nodes supply the other constraints. Adjacent cells therefore share height and first derivatives without forcing spatially distinct old samples to coincide.

Missing context is completed before connection, preserving all original valid values and derivatives. Missing regions use nearest-valid completion, with a finite constant fallback only for entirely missing tiles. Generated targets retain their actual finite mask. Height and normal-slope corrections are applied on the new side, with normalized corner weights to avoid summing multiple full-strength side corrections. The fixed variant uses a 16\,m band for both terms; slope taper keeps the height band fixed and shortens the derivative term's support, targeting at most 0.5\,m additional excursion with a minimum support of one sample spacing.

The joining experiment retains all 256 targets (64 terrain targets under four neighbor configurations). Large corrections receive quality diagnostics rather than being rejected from the comparison. The reported modification is
\begin{equation}
\Delta h_q=\left(\frac{1}{N_q}\sum_{i=1}^{N_q}
[h_q^+(x_i)-\widehat h_q(x_i)]^2\right)^{1/2}.
\label{eq:joining_modification}
\end{equation}
Here $N_q$ counts the fixed full-target samples. $H_c$ and $S_c$ measure height and slope disagreement through continuous queries at the preserved geometry boundary; they are distinct from the raw midpoint-extrapolated H/S in Table~\ref{tab:terrain_input_results}(b). Zeros in Table~\ref{tab:terrain_join_results}(c) denote residuals below $10^{-9}$. Their numerical closure follows from derivative inheritance, while $\Delta h_q$ quantifies the cost of modifying the candidate. Relative to fixed $C^1$ joining, slope taper reduces mean modification by 0.1263\,m (target-clustered 95\% bootstrap interval: 0.0829--0.1806\,m). The largest pointwise change remains 35.37\,m; exact boundary closure does not imply uniformly small modifications or gentle slopes throughout the connector.

\section{Hierarchical Scene Compilation}
\label{app:scene_details}

\subsection{Terrain Evidence and Spatial Responsibilities}

Evidence is computed before semantic allocation. For metric elevation $h$, the local gradient determines slope and the Laplacian summarizes curvature. After depression filling, drainage uses the steepest positive descent among adjacent cells:

\begin{equation}
\begin{aligned}
s(x)&=\|\nabla h(x)\|_2,\qquad k(x)=\nabla^2h(x),\\
d(x)&=\arg\max_{y\in\mathcal N_8(x)}
\frac{\widetilde h(x)-\widetilde h(y)}{\|x-y\|_2},
\qquad A(x)=1+\sum_{y:d(y)=x}A(y).
\end{aligned}
\label{eq:app_terrain_evidence}
\end{equation}

Here $\widetilde h$ is the depression-filled field; cells without positive descent have no receiver. Accumulation $A$ counts upstream contributing cells. The agent uses these fields together with inherited interfaces and asset capabilities to assign regional roles; evidence does not itself impose a semantic label.

Regions assign spatial responsibility, while districts refine it. Let $R_a$ be a region and $D_{ab}$ its districts on the target domain $\Omega_q$. Coverage and containment require

\begin{equation}
\bigcup_aR_a=\Omega_q,\qquad
D_{ab}\subseteq R_a,\qquad
\bigcup_bD_{ab}=R_a,\qquad
\operatorname{int}(D_{ab})\cap\operatorname{int}(D_{ac})=\emptyset\ (b\ne c).
\label{eq:app_region_partition}
\end{equation}

These conditions apply to ownership layers; roads, water corridors and vegetation overlays carry separate typed responsibilities. Boundary ports additionally constrain the position, direction and type of continuations. Region feasibility is checked before district and instance refinement, so later placement operates on accepted spatial responsibilities.

\subsection{Typed Relations and Asset Transforms}

An asset plan is a typed relation graph $\mathcal G_q=(\mathcal V_q^{\mathrm{asset}},\mathcal E_q^{\mathrm{rel}})$. Nodes carry catalog choice, role and district ownership; edges describe shared space, orientation, support or access. The catalog provides local geometry, support anchors, entrance axes and connector endpoints. For local point $p$ of instance $i$, the compiled transform is

\begin{equation}
p^{\,w}=A_i p+t_i,\qquad
T_i=\begin{bmatrix}A_i&t_i\\0&1\end{bmatrix},\qquad
\mathcal I_q=\{(\mathrm{id}_i,\mathrm{asset}_i,T_i,\mathrm{role}_i,\mathrm{district}_i)\}_i.
\label{eq:app_asset_transform}
\end{equation}

Allowed scaling is determined by asset type. For a two-ended connector module, let $p_-,p_+$ be local endpoints and $q_-,q_+$ the target endpoints. With $F(v)$ an orthonormal frame whose first axis follows $v$, alignment is

\begin{equation}
\begin{aligned}
\alpha&=\frac{\|q_+-q_-\|_2}{\|p_+-p_-\|_2},\\
A_i&=F(q_+-q_-)\operatorname{diag}(\alpha,1,1)F(p_+-p_-)^\top,\\
t_i&=(q_-+q_+)/2-A_i(p_-+p_+)/2 .
\end{aligned}
\label{eq:app_connector_transform}
\end{equation}

Only the longitudinal axis is scaled, and forbidden span changes are rejected. This transform aligns connector endpoints with their prescribed world positions. Other asset families use their own support and orientation contracts under the same typed interface.

\subsection{Geometric Validation and Repair Attribution}

The compiler evaluates the predicates required by each asset contract. With world support anchors $S_i$, admissible support surface $\mathcal S_i$, collision envelopes $B_i$, and access graph $\mathcal H$, representative constraints are

\begin{equation}
\begin{aligned}
 &\max_{p\in S_i}\operatorname{dist}(p,\mathcal S_i)\le\tau_i^{\mathrm{sup}},\\
 &\operatorname{dist}(B_i,B_j)\ge c_{ij}
 \quad\text{for pairs requiring clearance},\\
 &\operatorname{Reach}_{\mathcal H}(\mathrm{entry}_i,\mathrm{road}_i)=1
 \quad\text{for assets requiring road access}.
\end{aligned}
\label{eq:app_asset_checks}
\end{equation}

Support relations permit intended contact; clearance applies to incompatible object pairs. The tolerances and required predicates belong to the asset contract. Each failed predicate retains its instance, relation and district identifiers, which localize the plan entries to revise. Relation graphs therefore serve both construction and failure attribution, while deterministic geometry tools remain responsible for metric realization.

\section{Revision, Publication and Read-only Observation}
\label{app:query_details}

\subsection{Bounded Candidate Revision}

Program checks and visual review consume the same candidate revision. Let $g_j(\Delta M_q)$ be its required geometric predicates and $a_{\mathrm{reg}},a_{\mathrm{vis}}$ the region and compiled-scene review decisions. Publication requires every geometric predicate and both review decisions to pass:

\begin{equation}
\operatorname{Accept}(\Delta M_q)=
a_{\mathrm{reg}}\land a_{\mathrm{vis}}\land
\bigwedge_{j\in\mathcal J_{\mathrm{req}}}g_j(\Delta M_q).
\label{eq:app_acceptance}
\end{equation}

A revision delta identifies a set of editable candidate records $\mathcal U_t$. Unaffected semantic decisions remain fixed, while the compiler recomputes geometry that depends on edited records. The plan-level restriction and old-world protection are

\begin{equation}
\pi_q^{(t+1)}|_{\mathcal U_t^c}=\pi_q^{(t)}|_{\mathcal U_t^c},
\qquad
\operatorname{WriteSet}(\Delta M_q^{(t)})\cap\Omega_k=\emptyset.
\label{eq:app_revision_scope}
\end{equation}

A failed revision does not relax acceptance predicates. When the budget is exhausted, the candidate is rejected and the previous version remains available. Each revision regenerates reports and matched previews for its own candidate.

\subsection{Versioned Publication}

Publication binds terrain, instance transforms, plans and interface records into one version. Let $v$ be the visible version identifier and $\widehat M$ the validated candidate world. Readers select a version before querying:

\begin{equation}
\widehat M=M_k\oplus\Delta M_q,\qquad
v_{\mathrm{after}}=
\begin{cases}
k+1,&\operatorname{Accept}(\Delta M_q)=1\ \text{and publication succeeds},\\
k,&\text{otherwise}.
\end{cases}
\label{eq:app_publication}
\end{equation}

New records expose interfaces for subsequent additions, while existing instance identities and transforms persist. A query retains its selected version throughout a trajectory. Adding geometry may change visibility in a later version, but does not rewrite the earlier version's structural records.

\subsection{Camera Geometry and Video Conditions}

A viewing request provides a trajectory or lets the agent sample one from the requested motion and targets. The resulting camera schedule is $C_t=(K_t,R_t,c_t,\tau_t)$, where $R_t$ maps camera axes to world axes, $c_t$ is the camera center and $\tau_t$ is the timestamp. Clearance and requested visibility events are evaluated against the selected world. For homogeneous pixel $\bar u=(u,v,1)^\top$, the normalized world ray is

\begin{equation}
d_t(u)=\frac{R_tK_t^{-1}\bar u}{\|R_tK_t^{-1}\bar u\|_2},
\qquad
\lambda_t^*(u)=\min\{\lambda>0:c_t+\lambda d_t(u)\in\mathcal G(M_v)\}.
\label{eq:app_nearest_hit}
\end{equation}

The geometry set $\mathcal G(M_v)$ includes terrain and transformed assets. The nearest valid intersection determines visibility, hit identity and depth. Ray distance and camera-axis depth are distinct quantities:

\begin{equation}
X_t(u)=c_t+\lambda_t^*(u)d_t(u),\qquad
D_t^{\mathrm{ray}}(u)=\lambda_t^*(u),\qquad
D_t^z(u)=e_3^\top R_t^\top[X_t(u)-c_t].
\label{eq:app_depth_conventions}
\end{equation}

Here $e_3=(0,0,1)^\top$ is the camera's optical-axis unit vector. Pixels without a valid hit carry an explicit validity mask. The adapter preserves the depth convention and calibration while converting resolution, temporal sampling and encoding to the video model's interface. Depth conditions are paired with optional text or image rendering-style references supported by the selected generator. All queries and RGB synthesis operate on the selected world without a write-back path.
\subsection{Relation to Joint-Embedding Predictive Architectures}
\label{app:jepa}
I-JEPA learns semantic image representations by predicting target-region embeddings from context~\citep{ijepa2023}. V-JEPA extends feature prediction to video, learning representations through a latent prediction objective without pixel reconstruction~\citep{vjepa2024}. V-JEPA~2 further uses an action-conditioned predictor over learned representations for planning~\citep{vjepa22025}. These approaches support a broader distinction between internal world modeling and the synthesis of visual observations.

WorldWeave shares this motivation but implements the separation at the level of explicit world-state maintenance. Its persistent state comprises metric terrain, asset instances, semantic organization and cross-region interfaces. Hierarchical planning, geometric compilation and validated commits extend that state; a video generator receives depth queried from a selected version and optional appearance references. JEPA concerns learning predictive representations, whereas WorldWeave concerns constructing and preserving a geometrically addressable world for subsequent observation. The connection is conceptual rather than an adoption of the JEPA training objective: our contribution is the concrete state-construction, extension and read-only query mechanism.

\section{Additional Visual Comparisons}
\label{app:comparisons}
\begin{figure}[!htp]
\centering
\includegraphics[pagebox=cropbox,width=\linewidth]{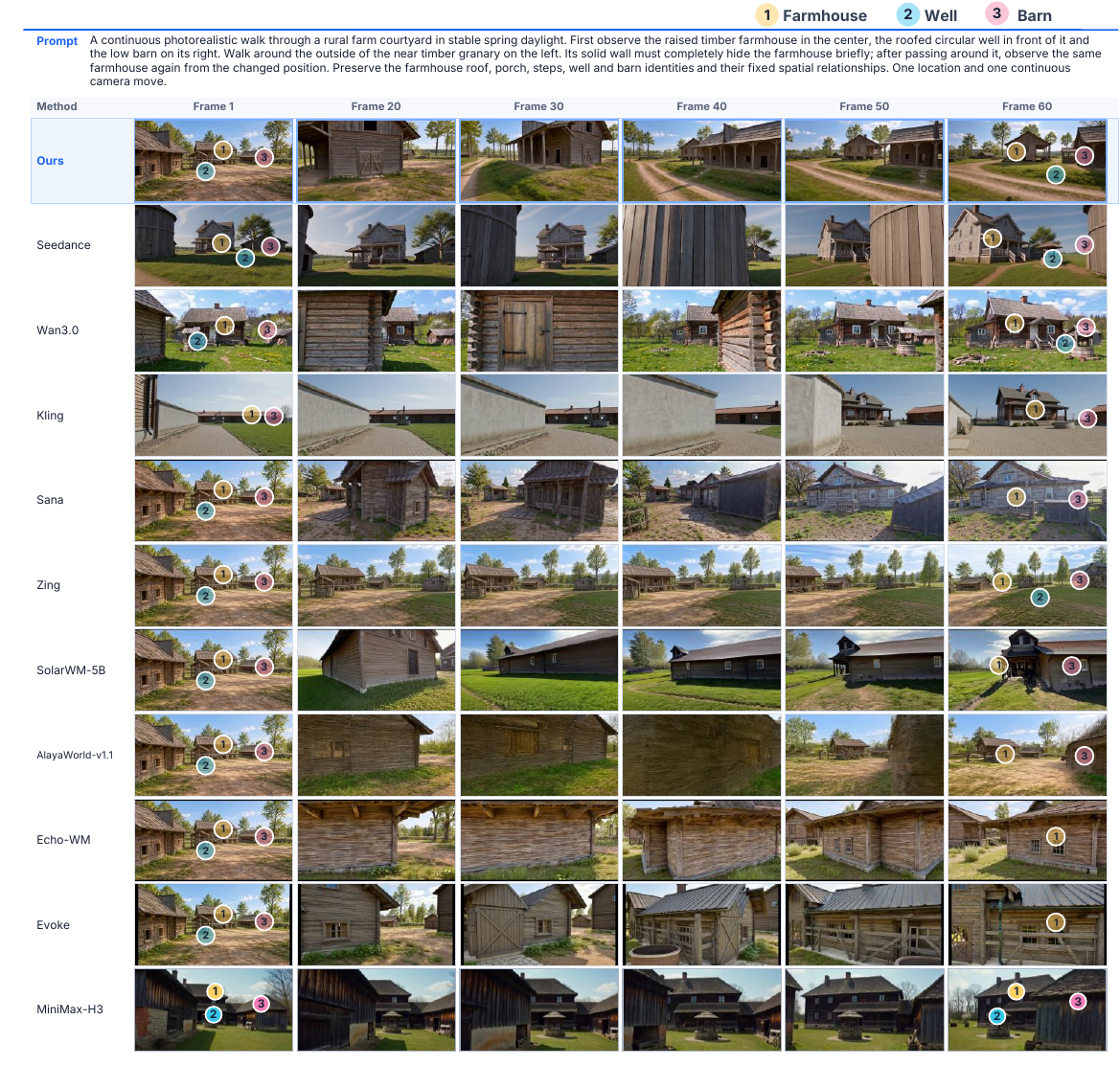}
\caption{Occlusion-and-return comparison in a farmstead. Rows show different methods and columns follow the video sequence. Numbered markers identify the farmhouse, well and barn for comparison across disappearance and return.}
\label{fig:appendix_compare1}
\end{figure}

\begin{figure}[!htp]
\centering
\includegraphics[pagebox=cropbox,width=\linewidth]{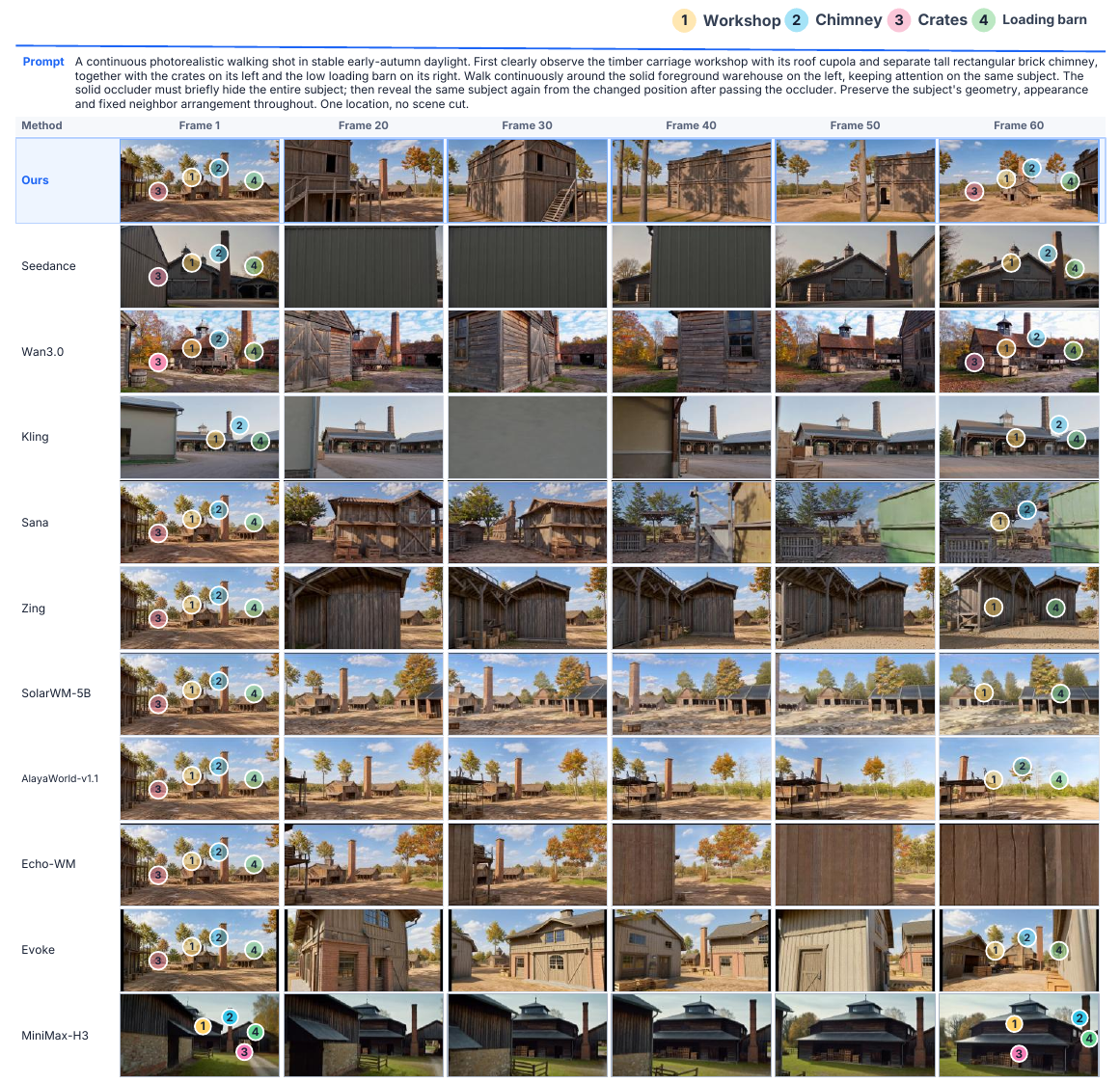}
\caption{Occlusion-and-return comparison in a town. Rows show different methods and columns follow the video sequence. Markers track the workshop, chimney, crates and loading barn as the camera passes behind the warehouse.}
\label{fig:appendix_compare2}
\end{figure}

\begin{figure}[!htp]
\centering
\includegraphics[pagebox=cropbox,width=\linewidth]{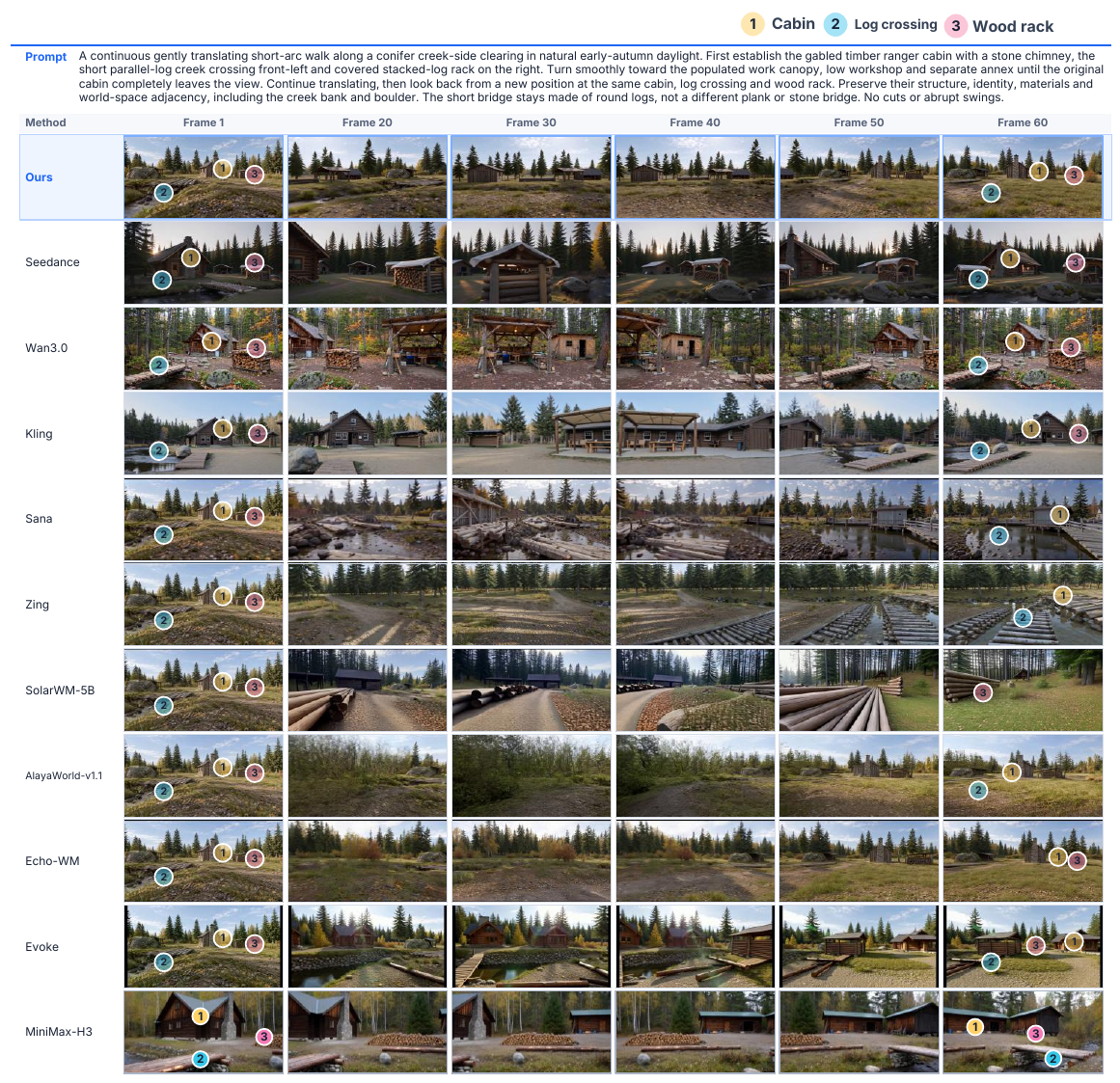}
\caption{Off-screen-return comparison in a woodland scene. Rows show different methods and columns follow the video sequence. Markers track the ranger cabin, log crossing and wood rack during the turn-away and return sequence.}
\label{fig:appendix_compare3}
\end{figure}

Figures~\ref{fig:appendix_compare1}, \ref{fig:appendix_compare2} and~\ref{fig:appendix_compare3} compare target persistence during occlusion and off-screen return in farmstead, town and woodland scenes. The sequences allow comparison of object identity, neighboring layout and completion of the requested camera motion. Each comparison includes all eleven methods, including MiniMax-H3 (Base).
\end{document}